\documentclass{article}

\usepackage{microtype}
\usepackage{graphicx}
\usepackage{subfigure}
\usepackage{booktabs} % for professional tables
\usepackage{amsmath,amssymb}
\usepackage{mathtools}
\usepackage{tikz}
\usetikzlibrary{positioning}

\usepackage{hyperref}

\usepackage[accepted]{icml2020}

\newcommand{\bcket}[3]{\left#1 #3 \right#2}
\renewcommand{\b}{\bcket{(}{)}}
\newcommand{\sqb}{\bcket{[}{]}}
\newcommand{\abs}{\bcket{\lvert}{\rvert}}
\newcommand{\ab}{\bcket{\langle}{\rangle}}
\renewcommand{\P}[1][]{\operatorname{P}_{#1}\b}

\newcommand{\N}{\mathcal{N}\b}
\newcommand{\h}{\mathbf{h}}
\renewcommand{\H}{\mathbf{H}}
\newcommand{\K}{\mathbf{K}}
\newcommand{\J}{\mathbf{J}}
\renewcommand{\L}{\mathbf{L}}
\newcommand{\R}{\mathbf{R}}
\newcommand{\T}{\mathbf{T}}

\newcommand{\U}{\mathbf{U}}
\newcommand{\A}{\mathbf{A}}
\newcommand{\M}{\mathbf{M}}

\newcommand{\V}{\mathbf{V}}
\newcommand{\W}{\mathbf{W}}
\newcommand{\I}{\mathbf{I}}
\newcommand{\Y}{\mathbf{Y}}
\newcommand{\X}{\mathbf{X}}
\newcommand{\y}{\mathbf{y}}

\renewcommand{\a}{\mathbf{a}}
\newcommand{\w}{\mathbf{w}}
\newcommand{\m}{\mathbf{m}}

\newcommand{\mS}{\mathbf{S}}
\renewcommand{\v}{\mathbf{v}}

\newcommand{\const}{\operatorname{const}}
\newcommand{\dd}[2][]{\frac{\partial #1}{\partial #2}}
\newcommand{\0}{{\bf {0}}}

\DeclareMathOperator*{\E}{\mathbb{E}}

\DeclareMathOperator*{\Cov}{\mathbb{C}}
\DeclareMathOperator*{\tr}{Tr}
\DeclareMathOperator*{\argmax}{arg\,max}

\icmltitlerunning{Why bigger is not always better: on finite and infinite neural networks}

\begin{document}

\twocolumn[
\icmltitle{Why bigger is not always better: on finite and infinite neural networks}
%\icmltitle{Submission and Formatting Instructions for \\
%           International Conference on Machine Learning (ICML 2020)}



\icmlsetsymbol{equal}{*}

\begin{icmlauthorlist}
\icmlauthor{Laurence Aitchison}{bris}
\end{icmlauthorlist}

\icmlaffiliation{bris}{University of Bristol, Bristol, UK}
%\icmlaffiliation{cam}{University of Cambridge, Cambridge, UK}

\icmlcorrespondingauthor{Laurence Aitchison}{laurence.aitchison@gmail.com}

\icmlkeywords{Bayesian neural network, Infinite neural network}

\vskip 0.3in
]



\printAffiliationsAndNotice{}  % leave blank if no need to mention equal contribution

\begin{abstract}
Recent work has argued that neural networks can be understood theoretically by taking the number of channels to infinity, at which point the outputs become Gaussian process (GP) distributed. However, we note that infinite Bayesian neural networks lack a key facet of the behaviour of real neural networks: the fixed kernel, determined only by network hyperparameters, implies that they cannot do any form of representation learning. The lack of representation or equivalently kernel learning leads to less flexibility and hence worse performance, giving a potential explanation for the inferior performance of infinite networks observed in the literature (e.g.\ Novak et al. 2019). We give analytic results characterising the prior over representations and representation learning in finite deep linear networks. We show empirically that the representations in SOTA architectures such as ResNets trained with SGD are much closer to those suggested by our deep linear results than by the corresponding infinite network. This motivates the introduction of a new class of network: infinite networks with bottlenecks, which inherit the theoretical tractability of infinite networks while at the same time allowing representation learning.
%We introduce infinite networks with bottlenecks and argue that as they do exhibit representation learning they are a better model for real networks.
%This confirms the need for a theoretical approach such as infinite networks with bottlenecks that can capture representation learning.
\end{abstract}

One approach to understanding and improving neural networks is to perform Bayesian inference in an infinitely wide network \citep{lee2018deep,matthews2018gaussian,garriga2019deep,novak2019bayesian}.
In this limit the outputs become Gaussian process distributed, enabling efficient and exact reasoning about uncertainty, and giving a means of interpretation using the parameter-free kernel function (which depends only on network hyperparameters such as depth).
However, the performance of Bayesian infinite networks lags considerably behind state-of-the-art finite networks trained using SGD (e.g. compare performance in \citet{garriga2019deep}, \citet{novak2019bayesian} and \citet{arora2019exact} against \citet{he2016identity} and \citet{chen2018neural}).
This seems surprising, because, to our knowledge, there are no reports of wider networks degrading classification performance \citep[indeed, the opposite is sometimes argued; see][]{zagoruyko2016wide}, and because exact Bayesian inference is provably optimal, if the prior accurately describes our beliefs \citep{ramsey2016truth}. %appears to be a very good strategy.  
Indeed, recent work on the Neural Tangent Kernel (NTK) \citep{li2019enhanced} has suggested that deterministic gradient descent in an infinite network gives slighly lower performance than Bayesian inference in the same network.

Our hypothesis is that the poor performance of Bayesian infinite networks arises because the top-layer representation (equivalent to the kernel), is fixed by the network hyperparameters, and thus cannot be learned from data.
This breaks many of our key intuitions about why deep networks are effective.
For instance in transfer learning \cite{huh2016makes} we use a large-scale dataset such as ImageNet to a learn a good high-level representation, then apply this representation to other tasks where less data is available.
However, transfer learning is impossible in infinite Bayesian neural networks, because the top-layer representation is fixed by the network hyperparameters and so cannot be learned using e.g.\ ImageNet.

To understand these issues, we analysed finite networks using tools from the infinite network literature \citep{lee2018deep,matthews2018gaussian,garriga2019deep,novak2019bayesian}.
We begin by giving a toy, two-layer example, contrasting the flexibility of finite networks with the inflexibility of infinite networks, showing that flexible finite networks offer benefits under conditions of model-mismatch.
We then introduce infinite networks with bottlenecks, which combine the theoretical tractability of infinite networks with the flexibility of finite networks.
To obtain an analytic understanding of kernel/representation flexibility and learning in such networks, we consider linear infinite networks with bottlenecks, which are equivalent to finite deep linear networks.
%
%We were able to show that, within the Bayesian setting, finite networks have the flexibility to learn useful high-level representations of data, while infinite networks do not.
%We begin by giving a toy, two-layer example, which demonstrates that finite networks offer potential benefits under conditions of model-mismatch.
We took two approaches to characterising these networks.
First, we considered the prior viewpoint, i.e.\ the covariance in the top-layer kernel induced by randomness in the lower-layer weights.
In particular, we showed that narrower, deeper networks offer more flexibility, and that CNNs offer more flexibility than locally connected networks (LCNs) when the input is spatially structured.
Second, we considered the posterior viewpoint, showing that under both MAP inference and posterior sampling, the representations in learned neural networks slowly transition from being similar to the input kernel (i.e. the inner product of the inputs) to being similar to the output kernel (i.e. the inner product of one-hot vectors representing the labels).
We found an important difference between MAP inference and sampling: for MAP inference, the learned representations transition from the input to the output kernel, irrespective of the network width.
Bayesian networks behave similarly when the network width and the number of output channels are equal, but 
%number of output channels has the same order as the network width, 
as the network width increases, the learned representations become increasingly dominated by the prior, and insensitive to the outputs.
%Finally, we considered representation learning in finite, deep, non-linear, convolutional networks, confirming the intuition that the networks learn a top-layer representation similar to the output kernel, and showing that the learned intermediate representations differ dramatically from those under the prior.
Remarkably, we find that in a ResNet trained using SGD on CIFAR-10, the representation differs dramatically from the corresponding infinite network and is instead very close to the output kernel, as suggested by our deep linear results.
This confirms the importance of working with a theoretical model, such as infinite networks with bottlenecks, that is capable of capturing representation learning.

\section{Toy Example}

\begin{figure*}[t]
  \centering
  \includegraphics{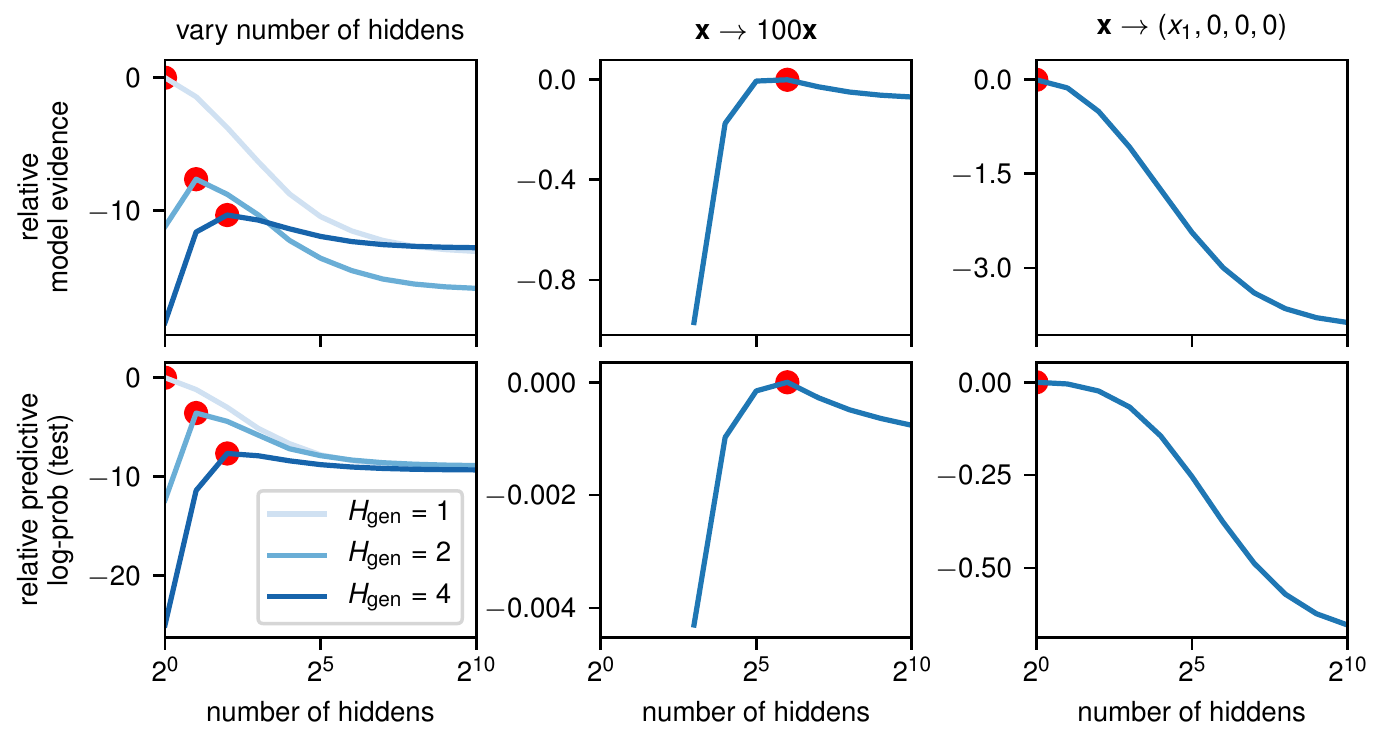}
  \caption{
    A toy fully-connected, two-layer Bayesian linear network showing situations in which smaller networks perform better than larger networks.
    The red dots indicate the optimal number of hidden units in that simulation.
    Left: training data generated from the prior with $H_\text{gen}$ hidden units.
    Middle: training data generated from the prior with $H_\text{gen}=4$ but where we scale-up the inputs by a factor of 100.
    Right: training data generated from the prior with $H_\text{gen}=4$, but where we zero-out all but the first input dimension.
    Top: Bayesian model evidence.
    Bottom: predictive log-probability, or equivalently test-error.
    \label{fig:toy}
  }
\end{figure*}
In the introduction, we noted that infinite Bayesian networks perform worse than standard neural networks trained using stochastic gradient descent.
Thus, as we make finite neural networks wider, there should be some point at which performance begins to degrade.
%This implies that as we make finite neural networks wider, there must be some point at which performance begins to degrade.
%Naively, however, this seems strange: to the best of our knowledge, there have been no reports of wider networks increasing classification error.
We considered a simple, two-layer, fully-connected linear network with the full set of $20$ 4-dimensional inputs denoted $\X$, hidden unit activations denoted $\H$, and 10-dimensional outputs denoted $\Y$,
\begin{align}
  \H &= \X \W 
  & \Y &= \H \V + \sigma \mathbf{\Xi}
\end{align}
where $\mathbf{\Xi}$ is IID standard Gaussian noise, $\W$ is the input-to-hidden weight matrix and $\V$ is the hidden-to-output weight matrix, whose columns, $\w_\mu$ and $\v_\nu$ are generated IID from,
\begin{align}
  \P{\w_\mu} &= \N{\w_\mu; \0, \tfrac{1}{X} \I} & \P{\v_\nu} &= \N{\v_\nu; \0, \tfrac{1}{H} \I},
\end{align}
and where the variance of the weights is normalised by the number of inputs to that layer, $X=4$ for the 4-dimensional input, and $H$ for the width of the hidden layer.

In the first example (Fig.~\ref{fig:toy} left), we generated targets for supervised learning using a second neural network with weights generated as described above, with $H_\text{gen} \in \{1, 2, 4\}$ hidden units. 
We evaluated the Bayesian model-evidence for networks with many different numbers of hidden units (x-axis).
Bayesian reasoning would suggest that the model evidence for the true model (i.e. with a matched number of hidden units) should be higher than the model evidence for any other model, as indeed we found (Fig.~\ref{fig:toy} top left), and these patterns held true for the predictive probability, or equivalently test performance (Fig.~\ref{fig:toy} bottom left).
While these results give an example where smaller networks perform better, they do not necessarily help us to understand the behaviour of neural networks on real datasets, where the true generative process for the data is not known, and is not in our model class.
As such, we considered two further examples where the neural network generating the targets lay outside of our model class.
In particular, we again generated target outputs by sampling a ``true'' network from the prior, but we modified the inputs to this network, first by multipling those inputs by $100$ (Fig.~\ref{fig:toy} middle), then by zeroing-out all but the first input unit (Fig.~\ref{fig:toy} right).
Critcally, we ensured model-mismatch by putting the original, unmodified inputs into the trained networks.
%First, we used the same neural network to generate the targets (with $H_\text{gen}=4$), but multiplying the inputs by $100$ . 
%Second, we modified the inputs by 
In both of these experiments, there was an optimium number of hidden units, after which performance degraded as more hidden units were included.

To understand why this might be the case, it is insightful to consider the methods we used to evaluate the model evidence and generate these results.
In particular, note that conditioned on $\H$, the output for any given channel, $\y_\nu$, is IID and depends only on the corresponding column of the output weights, $\v_\nu$,
\begin{align}
  %\P{\Y| \V, \H} &= \prod_\nu \P{\y_\nu| \v_\nu, \H} & \P{\y_\nu| \v_\nu, \H} &= \N{\y_\nu; \H \v_\nu, \sigma^2 \I}.
  \nonumber
  \P{\Y| \V, \H} &= \prod_\nu \P{\y_\nu| \v_\nu, \H} \\
                 &= \prod_\nu \N{\y_\nu; \H \v_\nu, \sigma^2 \I}.
  \intertext{Thus, we can integrate over the output weights, $\v_\nu$, to obtain a distribution over $\Y$ conditioned on $\H$,}
  \nonumber
  \P{\Y| \H} &= \prod_\nu \P{\y_\nu| \H} \\
  \label{eq:toy:gp}
             &= \prod_\nu \N{\y_\nu; \0, \tfrac{1}{H} \H \H^T + \sigma^2 \I}.
  %\P{\Y| \H} &= \prod_\nu \P{\y_\nu| \H} & \P{\y_\nu| \H} &= \N{\y_\nu; \0, \tfrac{1}{H} \H \H^T + \sigma^2 \I},
\end{align}
This is the classical Gaussian process representation of Bayesian linear regression \citep{williams2006gaussian}.
Remembering that the hidden activities, $\H$, is a deterministic function of the weights, $\W$, and inputs, $\X$, we can write this distribution as,
\begin{align}
  \nonumber
  \P{\y_\nu| \H} &= \P{\y_\nu| \W, \X} \\
                 &= \N{\y_\nu; \0, \tfrac{1}{H} \X \W \W^T \X^T + \sigma^2 \I}.
\end{align}
Thus, the first-layer weights, $\W$, act as kernel hyperparameters in a Gaussian process: they control the covariance of the outputs, $\y_\nu$.
To evaluate the model evidence we need to integrate over $\W$,
\begin{align}
  \nonumber
  \P{\Y| \X} &= \int \mathbf{dW} \P{\W} \prod_\nu \P{\y_\nu| \W, \X}\\
             &= \E_{\P{\W}}\sqb{ \prod_\nu \P{\y_\nu| \W, \X}}
\end{align}
and we estimate this integral by drawing $64\,000$ samples from the prior, $\P{\W}$.
Importantly, while $\W$ provides flexibility in the kernel in finite networks, this flexibility gradually disappears as we consider wider hidden layers networks.
In particular, 
\begin{align}
  \nonumber
  \lim_{H \rightarrow \infty} \tfrac{1}{H} \W \W^T 
  &= \lim_{H \rightarrow \infty} \tfrac{1}{H} \sum_{\mu=1}^H \w_\mu \w_\mu^T \\
  &= \E\sqb{\w_\mu \w_\mu^T} = \tfrac{1}{X} \I.
\end{align}
Therefore, in this limit, the distribution over $\Y$ converges to,
\begin{align}
  \lim_{H \rightarrow \infty} \P{\Y| \X} &= \prod_\nu \N{\y_\nu; \0, \tfrac{1}{X} \X \X^T + \sigma^2 \I}.
\end{align}
This is exactly the distribution we would expect from Bayesian linear regression in a one-layer network.
Thus, by taking the infinite limit, we have eliminated the additional flexibility afforded by the two-layer network, and we can see that the superior performance of smaller networks in Fig.~\ref{fig:toy} emerges because they give additional flexibility in the covariance of the outputs, which gradually disappears as network size increases.
Finally, note that sampling from the prior works well here both because of the concentration result above, and because we use relatively small amount of data, 20 points.

%In the rest of the paper, we elaborate on this fundamental intuition that finite neural networks allow for flexibility in the covariance of the outputs that is eliminated as we move to infinite networks. 
%We begin by showing that the kernel representation, (i.e. $\X^T \X/X$ or $\H^T \H/H$) is complete in the sense that it entirely determines the prior distribution over activities at the next layer.
%We then assess the convergence of the output kernel, $\H^T \H / H$ as a function of the number of hidden units, finding that the standard deviation of an element of the kernel has the standard $1/\sqrt{H}$ scaling.
%Next, we consider the specific values of this variance for linear and nonlinear (rectified) layers.

\section{Infinite networks with finite bottlenecks}
\newcommand{\myoverbrace}[2]{{\color{gray} \overbrace{\color{black} #1}^{#2}}}
\newcommand{\myunderbrace}[2]{{\color{gray} \underbrace{\color{black} #1}_{#2}}}
\begin{figure*}[t]
  \centering
  \begin{tikzpicture}
    \def\d{5cm}
    \node (x) at ({-0.7*\d}, 0) {$\myoverbrace{\H_0}{P \times M_0} = \X$};
    \node (a) at (0, 0) {$\myoverbrace{\A_\ell}{P \times N_\ell} = \H_{\ell-1} \W_\ell$};
    \node (ap) at ({\d}, 0) {$\myoverbrace{\A_\ell'}{P\times M_\ell} = \A_\ell \M_\ell$};
    \node (h) at ({2*\d}, 0) {$\myoverbrace{\H_\ell}{P\times M_\ell} = \phi(\A'_\ell)$};
    \node[anchor=north] (L0) at (x.south) {$\myunderbrace{\L_0 =\tfrac{1}{M_0} \X \X^T}{\text{input kernel}}$};
    \node[anchor=north] (J) at (a.south) {$\myunderbrace{\J_\ell = \Cov\sqb{\a_\mu^\ell| \H_{\ell-1}}}{\text{covariance}}$};
    \node[anchor=north] (K) at (ap.south) {$\myunderbrace{\K_\ell = \tfrac{1}{N_\ell} \A_\ell \A_\ell^T = \tfrac{1}{M_\ell} \A_\ell' \A_\ell'^T}{\text{activation kernel}}$};
    \node[anchor=north] (L) at (h.south) {$\myunderbrace{\L_\ell = \tfrac{1}{M_\ell} \H_\ell \H_\ell^T}{\text{activity kernel}}$};
    \draw[->] (x) to (a);
    \draw[->] (a) to (ap);
    \draw[->] (ap) to (h);
    \draw[->] (h) .. controls +(90:1.5cm) and +(0:2cm) .. 
    (\d,1.5cm) .. controls +(180:2cm) and +(90:1.5cm) .. (a);
  \end{tikzpicture}
  \caption{
    The relationships between the feature-space and kernel representations of the neural network.
    For a typical finite neural network, $\M_\ell = \I$, so $M_\ell = N_\ell$.
    For a finite-infinite network (which allows us to compute $\L_\ell$ from $\K_\ell$), we send $M_\ell \rightarrow \infty$, and draw the elements of $\M_\ell$ IID from a Gaussian distribution with zero mean and variance $1/M_\ell$.
    \label{fig:schem}
  }
\end{figure*}
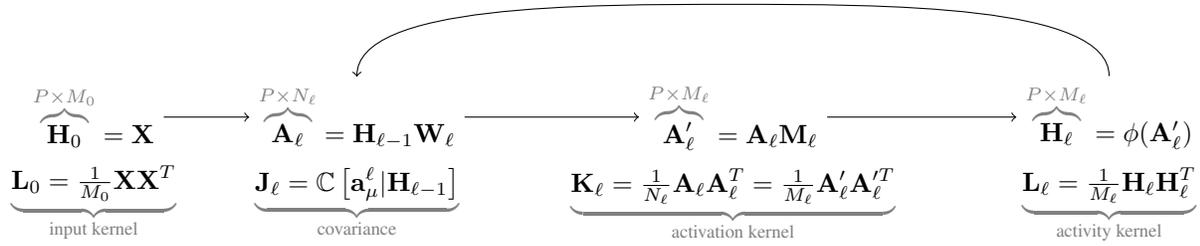
In the previous section, we considered the simplest networks in which these phenomena emerge: a two-layer, linear network.
In this section, we setup a full infinite network with bottlenecks and show that activity flowing through this network can be understood entirely in terms of kernel and covariance matricies.

%Here we show that the same phenomenon --- the reduction in flexibility in the kernel as we widen the network --- occurs in deep, non-linear networks.
%Along the way, we obtain two remarkable results.
%First, we find that outputs from a layer depend only on the kernel defined by the inputs to that layer, and not on the full details of those inputs --- this corresponds to a fundamental rotational symmetry at the heart of all neural networks.
%Second, we find that output channels a layer are IID conditioned on the input kernel, which gives a strong tool for analysing the convergence of the kernel.
%As such, if this is the last layer, we can readily characterise the output distribution, and if this is an intermediate layer, we can readily analyse the convergence of the kernel defined by the outputs to its expected value.

Consider a single layer within a fully-connected network, where the potentially infinite activity at the previous layer, $\H_{\ell-1}$, corresponding to a batch containing all inputs, is multiplied by a weight matrix, $\W_\ell$, to give a finite number of activations, $\A_\ell$. 
This activation matrix, $\A_\ell$ is multiplied by another matrix, $\M_\ell$, to give a potentially infinite updated activation matrix, $\A_\ell'$, which is then passed through a non-linearity, $\phi$, to give the potentially infinite activity at this layer, $\H_{\ell}$.
Note that following \citet{matthews2018gaussian}, we use ``activation'' pre-nonlinearity and ``activity'' post-nonlinearity.
%Critically, this multiplication leaves the representation unchanged (Fig.~\ref{fig:schem}): $\M_\ell$ ensures that the kernel for the output of the output of the nonlinearity can be computed.
%Finally, the activations, $\A_\ell'$, are passed through a non-linearity, $\phi$, 
\begin{align}
  \label{eq:def:HA}
  \A_\ell &= \H_{\ell-1} \W_\ell & \A_\ell' &= \A_\ell \M_\ell & \H_\ell &= \phi\b{\A_\ell'}
  \intertext{where the input data is $\H_0 = \X$, and}
%\end{align}
%where,
%\begin{align}
  \nonumber
  \H_\ell &\in \mathbb{R}^{P \times M_\ell} &
  \A_\ell &\in \mathbb{R}^{P \times N_\ell} &
  \A_\ell' &\in \mathbb{R}^{P \times M_\ell} \\
  \W_\ell &\in \mathbb{R}^{M_{\ell-1} \times N_\ell} &
  \M_\ell &\in \mathbb{R}^{N_\ell \times M_\ell}
\end{align}
For an infinite network with bottlenecks, we take the limit as $M_\ell$ goes to infinity, leaving $N_\ell$ finite.
As such, the activity before, $\A_\ell'$, and after, $\H_\ell$, the nonlinearity is infinite, with a finite linear bottleneck formed by $\A_\ell$.

For a fully-connected network, the columns of $\W_\ell$ and $\M_\ell$, denoted $\w^\ell_\lambda$ and $\m^\ell_\lambda$ are generated IID from a Gaussian distribution,
\begin{align}
  \label{eq:def:wmu_hmu}
  \P{\W_\ell} &= \prod_{\lambda=1}^{N_\ell} \P{\w_\lambda^\ell} = \prod_{\lambda=1}^{N_\ell} \N{\w_\lambda^\ell; \0, \tfrac{1}{M_{\ell-1}} \I}\\ %& \P{\M_\ell} &= \prod_{\mu=1}^{N_\ell} \N{\m_\mu^\ell; \0, \tfrac{1}{N_{\ell}} \I},
  \P{\M_\ell} &= \prod_{\lambda=1}^{M_\ell} \P{\m^\ell_\lambda} = \prod_{\lambda=1}^{M^\ell_\lambda} \N{\m^\ell_\lambda; \0, \tfrac{1}{N_\ell} \I}.
\end{align}
where the normalization constants, $1/M_{\ell-1}$ and $1/N_{\ell-1}$, ensure that activations remain normalized as they flow through the network.
%In a deep finite linear network, we would set $\M_\ell = \I$, such that $\A_\ell' = \A_\ell$ and $M_\ell = N_\ell$.
%But in the usual infinite network with finite bottleneck setting, 
%\begin{align}
%  \P{\M_\ell} &= \prod_{\lambda=1}^{M_\ell} \P{\m^\ell_\lambda} = \prod_{\lambda=1}^{M^\ell_\lambda} \N{\m^\ell_\lambda; \0, \tfrac{1}{N_\ell} \I}.
%\end{align}

Following the infinite network literature, we would like to characterise activity flowing through the network in terms of the activation kernel, $\K_\ell$ and activity kernel, $\L_\ell$,
\begin{align}
  \label{eq:def:KL}
  \nonumber
  \K_{\ell} &\equiv \tfrac{1}{N_{\ell}}\A_{\ell} \A_{\ell}^T \\%= \tfrac{1}{M_\ell} \A_\ell'\A_\ell'^T \\
  \L_{\ell} &\equiv \tfrac{1}{M_{\ell}}\H_{\ell} \H_{\ell}^T \quad\quad\quad\quad\quad \L_0 = \tfrac{1}{M_0}\X \X^T 
\end{align}

We begin by characterising the relationship between $\A'_\ell$ and $\A_\ell$.
As each channel (column) of $\A'_\ell$ is a linear function of the corresponding channel of the weights, $\a'^\ell_\lambda = \A_\ell \m_\lambda^\ell$, these activations are Gaussian and IID conditioned on $\A_\ell$,
%By analogy with Eq.~\eqref{eq:def:P(A|H)}, $\A'_\ell$ is Gaussian distributed conditioned on $\K_\ell=\tfrac{1}{N_\ell} \A_\ell \A_\ell^T$,
\begin{align}
  \nonumber
  \P{\A'_\ell| \A_\ell} &= \prod_{\lambda=1}^{M_\ell} \P{\a'^{\ell}_\lambda| \A_\ell} \\
  &= \prod_{\lambda=1}^{M_\ell} \N{\a'^{\ell}_\lambda; \0, \K_\ell} = \P{\A'_{\ell}| \K_\ell}
\end{align}
and taking the limit of $M_\ell \rightarrow \infty$,
\begin{align}
  \nonumber
  \lim_{M_\ell \rightarrow \infty} \tfrac{1}{M_\ell} \M_\ell \M_\ell^T &= \tfrac{1}{N_\ell} \I \\ 
  \nonumber
  \lim_{M_\ell \rightarrow \infty} \tfrac{1}{M_\ell} \A'_\ell \b{\A'_\ell}^T &= \lim_{M_\ell \rightarrow \infty} \tfrac{1}{M_\ell} \A_\ell \M_\ell \M_\ell^T \A_\ell^T \\
  &= \tfrac{1}{N_\ell} \A_\ell \A_\ell^T = \K_\ell,
\end{align}
Thus, the kernel for $\A_\ell$ is equivalent to the kernel for $\A'_\ell$ in infinite networks with finite bottlenecks (Fig.~\ref{fig:schem}).

Next, consider computing $\K_\ell$ from $\L_{\ell-1}$.
As each channel (column) of the activations is a linear function of the corresponding channel of the weights, $\a^\ell_\lambda = \H_{\ell-1} \w^\ell_\lambda$, the activations are Gaussian and IID conditioned on the activity at the previous layer,
\begin{align}
  \label{eq:def:P(A|H)}
  \nonumber
  \P{\A_\ell| \H_{\ell-1}} &= \prod_{\mu=1}^{N_\ell} \P{\a^\ell_\lambda| \H_{\ell-1}}\\
                           &= \prod_{\mu=1}^{N_\ell} \N{\a^\ell_\lambda; \0, \J_\ell}
                           = \P{\A_\ell| \J_\ell},
\end{align}
with covariance $\J_\ell$.
For a fully connected network, the covariance, $\J_\ell$, is equal to the previous layer's activity-kernel, $\L_\ell$,
\begin{align}
  \label{eq:def:J}
  \J_\ell &= \L_{\ell-1} = \tfrac{1}{M_{\ell-1}}\H_{\ell-1} \H_{\ell-1}^T,
\end{align}
but the relationship is more complex in convolutional architectures \citep{garriga2019deep,novak2019bayesian} (Appendix~\ref{app:prior:conv}).
As $\A_\ell$ is always finite and random, $\K_\ell$ is also a random variable, and inspecting the above expressions, its distribution can be written as a Wishart, centered on $\L_{\ell-1}$.

Finally consider computing $\L_\ell$ from $\K_\ell$.
Note that as both $\A'_\ell$ and $\H_\ell$ are infinite, we can directly use standard results from infinite neural networks, i.e. those from \citet{cho2009kernel}, as in \citet{lee2018deep,matthews2018gaussian,garriga2019deep,novak2019bayesian}.

%\subsection{Deep linear networks}

%Critically,

%Thus, we can directly wriite the distribution over $\K_\ell$ given $\L_{\ell-1}$ as a Wishart, or it can be left implicit. 

%By analogy with Eq.~\eqref{eq:def:P(A|H)}, $\A'_\ell$ is Gaussian distributed conditioned on $\K_\ell=\tfrac{1}{N_\ell} \A_\ell \A_\ell^T$,
%\begin{align}
%  \nonumber
%  \P{\A'_\ell| \A_\ell} &= \prod_{\lambda=1}^{M_\ell} \P{\a'^{\ell}_\lambda| \A_\ell} \\
%  &= \prod_{\lambda=1}^{M_\ell} \N{\a'^{\ell}_\lambda; \0, \K_\ell} = \P{\A'_{\ell}| \K_\ell}
%\end{align}
%and taking the limit of $M_\ell \rightarrow \infty$,
%\begin{align}
%  \nonumber
%  \lim_{M_\ell \rightarrow \infty} \tfrac{1}{M_\ell} \M_\ell \M_\ell^T &= \tfrac{1}{N_\ell} \I \\ 
%  \nonumber
%  \lim_{M_\ell \rightarrow \infty} \tfrac{1}{M_\ell} \A'_\ell \b{\A'_\ell}^T &= \lim_{M_\ell \rightarrow \infty} \tfrac{1}{M_\ell} \A_\ell \M_\ell \M_\ell^T \A_\ell^T \\
%  &= \tfrac{1}{N_\ell} \A_\ell \A_\ell^T = \K_\ell,
%\end{align}
%then we simultaneously have Gaussian $\P{\A_\ell'| \K_\ell}$, and we have left the activation kernel unchanged.
%As this network alternates between finite $N_\ell$ and infinite $M_\ell$ layers, we call it a finite-infinite network, but work published just after ours these networks have also been called ``infinite networks with finite bottlenecks'' \citep{agrawal2020wide}.

Linear infinite networks with finite bottlenecks can be obtained by setting $\H_\ell = \phi(\A'_\ell) = \A'_\ell$, implying that $\L_\ell = \K_\ell$.
Critically, this is equivalent to a deep linear network obtained by in adddition setting $\M_\ell = \I$ so that $\A_\ell' = \A_\ell$ and $M_\ell = N_\ell$, as these choices imply that the $\A_\ell = \A_\ell' = \H_\ell$ so that again, $\L_\ell = \K_\ell$.

\subsection{DNNs are deep GPs}

Given this setup, we can see that even a finite nonlinear network (i.e.\ with $\M_\ell = \I$) is a deep Gaussian process.
In particular, in a deep Gaussian process, the activations at layer $\ell$, denoted $\A_{\ell}$, consist of $N_{\ell}$ IID channels that are Gaussian-process distributed (Eq.~\ref{eq:def:P(A|H)}), with a kernel/covariance determined by the activations at the previous layer.
For a fully connected network,
\begin{align}
  \J_\ell &= \L_{\ell-1} = \tfrac{1}{N_{\ell-1}} \phi\b{\A_{\ell-1}}\phi^T\b{\A_{\ell-1}}.
\end{align}
The relationship between finite neural networks and deep GPs is worth noting, because the same intuition, of the lower-layers shaping the top-layer kernel, arises in both senarios \citep[e.g.][]{bui2016deep}, and because there is potential for applying GP inference methods for neural networks, and vice versa.

\section{The prior view on kernel flexibility}
We can analyse how flexibility in the kernel emerges by looking at the variability (i.e.\ the variance and covariance) of $\J_\ell$, $\K_\ell$ and $\L_\ell$.
If the prior gives a stochastic kernel with higher variance, then it will be easier to shape that kernel by conditioning on data.
In the appendix, we derive recursive updates for deep, linear, convolutional networks, but here, for simplicity we give the fully-connected updates,
\begin{subequations}
\label{eq:prior}
\begin{align}
  \label{eq:CovJ}
  \Cov\sqb{J^{\ell}_{ij}, J^{\ell}_{kl}| \L^0} =& \Cov\sqb{L^{\ell-1}_{ij}, L^{\ell-1}_{kl} |\K^0}\\
  \label{eq:CovK}
  \Cov\sqb{K^{\ell}_{ij}, K^{\ell}_{kl}| \L^0} \approx&
    \Cov\sqb{J^{\ell}_{ij}, J^{\ell}_{kl}| \L^0} \\
  \nonumber
    &+\tfrac{1}{N_\ell} \b{\ab{J^\ell_{ik}} \ab{J^\ell_{jl}} + \ab{J^\ell_{il}} \ab{J^\ell_{jk}}}\\
  \Cov\sqb{L^{\ell}_{ij}, L^{\ell}_{kl}| \L^0} =& 
    \Cov\sqb{K^{\ell}_{ij}, K^{\ell}_{kl}| \L^0}
  \intertext{where,}
  \ab{J^\ell_{ij}} &= \E\sqb{J^\ell_{ij}| \L^0} = L_{ij}^0
\end{align}
\end{subequations}
%
%\begin{subequations}
%\label{eq:prior}
%\begin{align}
%  \label{eq:CovJ}
%  \Cov\sqb{J^{\ell}_{ir,jr}, J^{\ell}_{ku,lu}| \L^0} =& 
%  \tfrac{1}{D_{\ell-1}^2} \sum_{dd'} \Cov\sqb{L^{\ell-1}_{i(r+d),j(r+d)}, L^{\ell-1}_{k(u+d'),l(u+d')}| \L^0}\\
%  \label{eq:CovK}
%  \nonumber
%  \Cov\sqb{K^{\ell}_{ir,jr}, K^{\ell}_{ku,lu}| \L^0} \approx&
%    \Cov\sqb{J^{\ell}_{ir,jr}, J^{\ell}_{ku,lu}| \L^0} \\
%    &+\tfrac{1}{N_\ell} \b{\ab{J^\ell_{ir,ku}} \ab{J^\ell_{jr,lu}} + \ab{J^\ell_{ir,lu}} \ab{J^\ell_{jr,ku}}} + \mathcal{O}(1/N_\ell^2)\\
%  \Cov\sqb{L^{\ell}_{ir,jr}, L^{\ell}_{ku,lu}| \L^0} =& 
%    \Cov\sqb{K^{\ell}_{ir,jr}, K^{\ell}_{ku,lu}| \L^0}
%  \intertext{where,}
%  \ab{J^\ell_{ir,jr}} &= \E\sqb{J^\ell_{ir,jr}| \L^0}
%\end{align}
%\end{subequations}
where $i$, $j$, $k$ and $l$ index datapoints.

This expression predicts that the variance of the kernel is proportional to the depth (including the last layer; $L+1$) and inversely proportional to the width, $N$,
\begin{align}
  \Cov\sqb{K^{L+1}_{ij}, K^{L+1}_{kl}| \L^0} \approx \tfrac{L+1}{N} \b{L_{ik}^0 L_{jl}^0 + L_{il}^0 L_{jk}^0}.
\end{align}
This expression is so simple because, for a fully-connected linear network, the expected covariance at each layer is the same.
For nonlinear and convolutional or locally-connected networks the covariance is still proportional to $1/N$, but the depth-dependence becomes more complex, as the covariance changes as it propagates through layers.

%\subsection{Kernel flexibility: Prior viewpoint: Experimental}
\begin{figure*}[t]
  \centering
  \includegraphics{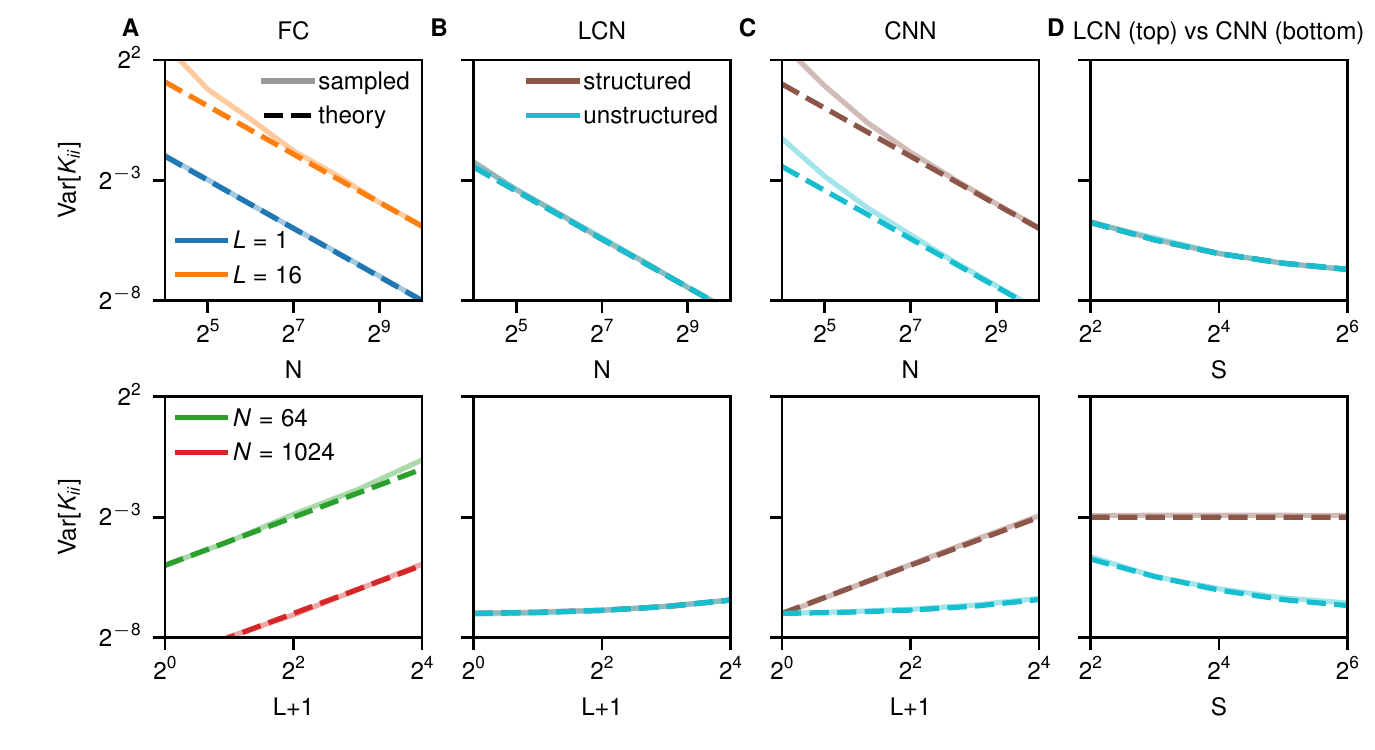}
  \caption{
    The variance of the stochastic kernel induced by randomly sampling weights in finite, linear, fully connected and convolutional networks, with spatially structured and unstructured inputs.
    We use normalized inputs and circular convolutions to ensure that the kernel's expected value remains equal to $1$ at all locations as it propagates through the network.
    The dashed lines in all plots display the theoretical approximation (Eq.~\ref{eq:prior}) which is valid when the width is much greater than the number of layers.
    The solid lines display the empirical variance of the kernel from 10,000 simulations.
    \textbf{A} The variance of the kernel for fully connected networks, plotted against network width, $N$, for shallow (blue; $L+1=1$), and deep (orange; $L+1=16$) networks (top) and plotted against network depth, $L+1$, for narrow (green; $N=64$) and wide (red; $N=1024$) networks.
    \textbf{B} The variance of the kernel for locally connected networks with spatially structured and unstructured inputs, plotted against the number of channels, $N$, and against network depth, $L+1$.  Note that the structured line lies underneath the unstructured line.
    The inputs are 1-dimensional with $S=32$ spatial locations, and $100$ input channels.
    \textbf{C} As in \textbf{B}, but for convolutional networks.
    \textbf{D} The variance of the kernel as a function of the input spatial size, $S$, for deep ($L+1=16$) LCNs (top) and CNNs (bottom) with spatially strutured and unstructured inputs.
    \label{fig:prior}
  }
\end{figure*}
To check the validity of these expressions, we sampled 10,000 neural networks from the prior, and evaluated the variance of the kernel for a single input (Fig.~\ref{fig:prior}).
These inputs were either spatially unstructured (i.e.\ white noise), or spatially structured, in which case the inputs were the same across the whole image.
For fully connected networks, we confirmed that the variance of the kernel is proportional to the depth including the last layer, $L+1$, and inversely proportional to width, $N$ (Fig.~\ref{fig:prior}A).
For locally connected networks, we found that structured and unstructured inputs gave the same kernel variance, which is expected as any spatial structured is destroyed after the first layer (Fig.~\ref{fig:prior}B).
Further, for convolutional networks with structured input, the variance of the kernel was proportional to network depth (Fig.~\ref{fig:prior}C bottom), but whenever that spatial structure was absent, either because it was absent in the inputs or because it was eliminated by an LCN (Fig.~\ref{fig:prior}BC bottom) the variance of the kernel was almost constant with depth (see Appendix~\ref{sec:app:space}).

The large decrease in kernel flexibility for locally connected networks might be one reason behind the result in \citet{novak2019bayesian} that locally connected networks have performance that is very similar to an infinite-width network, in which all flexibility has been eliminated.
In essence, for a locally connected network, we sample the weights for each spatial region independently, so we in effect average over more IID random variables, reducing the variance of the kernel at the next layer, and hence reducing the possibility for data to shape that representation.
In contrast, for a convolutional network we share weights across locations, increasing the variance in the kernel, and hence increasing the possibility for data to shape the representation.
%convolutional network, we share weights across locally-connected regions, 
Finally, as the spatial input size, $S$, increases, for convolutional networks with spatially structured inputs, the variance of the kernel is constant, whereas for locally connected or spatially unstructured inputs, the variance falls (Fig.~\ref{fig:prior}D).

\section{The posterior view on kernel flexibility}
An alternative approach to understanding flexibility in finite neural networks is to consider the posterior viewpoint: how learning shapes top-level representations.
To obtain analytical insights, we considered maximum a-posteriori and sampling based inference in a deep, fully-connected, linear network.
In both cases, we found that learned neural networks shift the representation from being close to the input kernel, defined by,
\begin{align}
  \K_0 = \L_0 = \tfrac{1}{M_0} \X \X^T,
\end{align}
to being close the output kernel, defined by,
\begin{align}
  \K_{L+1} = \tfrac{1}{N_{L+1}} \Y \Y^T.
\end{align}

In particular, under MAP inference, the shape of the kernel smoothly transitions from the input to the output kernel (Appendix~\ref{app:post:map}),
\begin{align}
  \K_\ell &= \b{\frac{N_{\ell<}}{N_{\leq \ell}}}^{\tfrac{\ell (L+1-\ell)}{L+1}} \b{\K_{L+1} \K_0^{-1}}^{\ell/(L+1)}  \K_0,
\end{align}
where $N_{\ell<}$ is the geometric average of the width in layers $\ell+1$ to $L+1$, and $N_{\leq\ell}$ is the geometric average of the width in layers $1$ to $\ell$.
Thus, the kernels (and the underlying weights) at each layer can be made arbitrarily large or small by changing the width, despite the prior distribution being chosen specifically to ensure that the scale of the kernels was invariant to network width.
This is an issue inherent to the use of MAP inference, which often finds modes that give a poor characterisation of the Bayesian posterior.
In contrast, if we sample the weights using Langevin sampling (Appendix~\ref{app:post:langevin}), and set all the intermediate widths, from $N_1$ to $N_L$ to $N$, then we get a similar expression,
\begin{align}
  \K_\ell &= \b{\K_L \K_0^{-1}}^{\ell/L}  \K_0,
\end{align}
where the kernels slowly transition from $\K_0$ to $\K_L$.
The key difference is that the similarity between the top-layer representation, $\K_L$, and the output kernel, $\K_{L+1}$, depends on the ratio between the network width, $N$, and the number of output units, $Y=N_{L+1}$.
In particular, if $Y=N$, then we get a relationship very similar to that for MAP inference,
\begin{align}
  \K_\ell &= \b{\K_{L+1} \K_0^{-1}}^{\ell/(L+1)}  \K_0,
\end{align}
%as $\K_{L} = \b{\K_{L+1} \K_0^{-1}}^{L/(L+1)} \K_0$.
However, as the network width grows very large, the prior begins to dominate, and the posterior becomes dominated by the prior,
\begin{align}
  \lim_{N/Y \rightarrow \infty} \K_\ell &= \K_0,
\end{align}
as $\K_L = \K_0$.
Finally, if the network width is small in comparison to the number of units,
\begin{align}
  \lim_{N/Y \rightarrow 0} \K_\ell &= \b{\K_{L+1} \K_0^{-1}}^{\ell/L}  \K_0,
\end{align}
as the top-layer kernel converges to the output, $\K_{L} = \K_{L+1}$.

\begin{figure*}[t]
  \centering
  \includegraphics{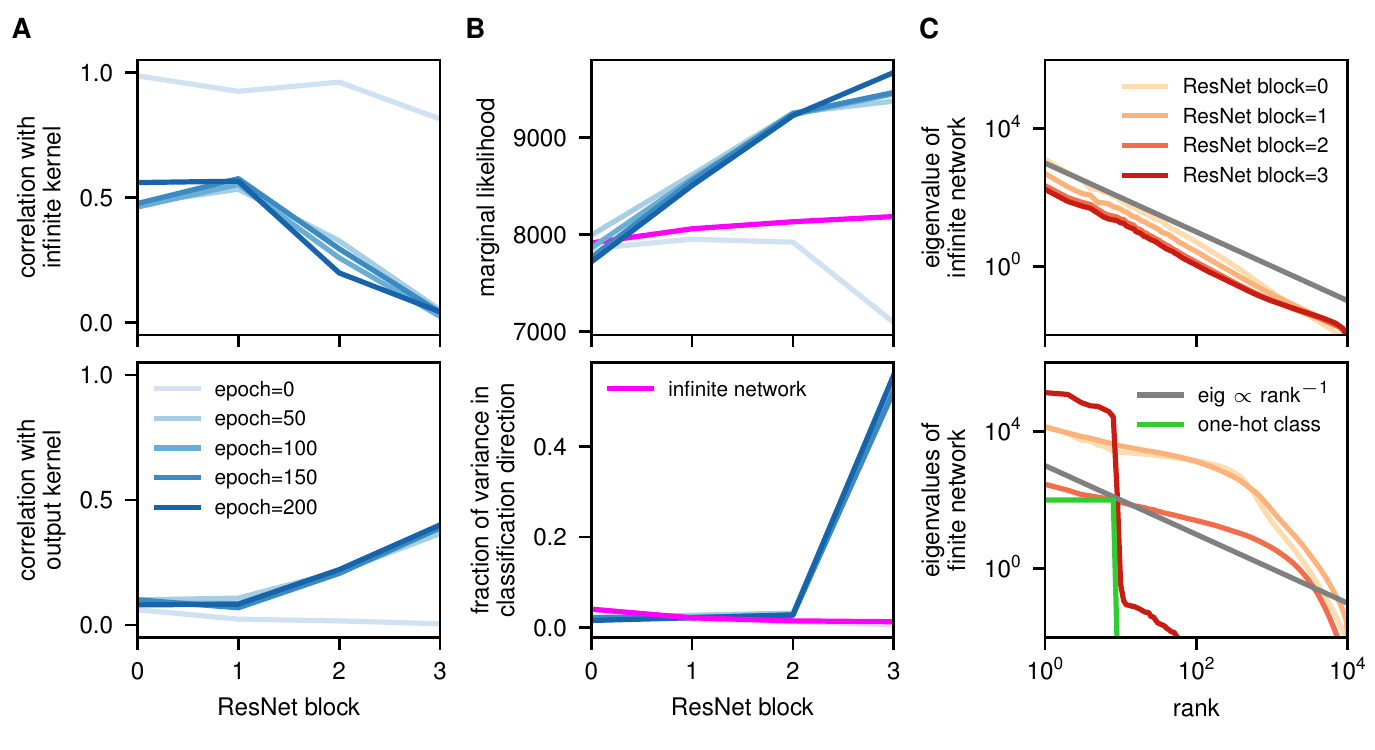}
  \caption{
    Comparison of kernels for finite and infinite neural networks at different layers.  All kernels are computed on test data.
    \textbf{A} (top) Correlation (coefficient) between the kernel defined by the infinite network, and kernel defined by a finite network after different numbers of training epochs.
    \textbf{A} (bottom) Correlation (coefficient) between the kernel defined by the infinite network, and the output kernel defined by taking the inner product of one-hot vectors representing the class label.
    \textbf{B} (top) The Gaussian process marginal likelihood for the 10 functions given by the one-hot class labels, evaluated using the kernel output by different ResNet blocks.
    \textbf{B} (bottom) The fraction of variance in the direction of the one-hot output class labels.
    \textbf{C} (top) The eigenvalues of the kernel defined by the infinite network as we progress through layers, and compared to a $-1$ power law (grey).
    \textbf{C} (top) The eigenvalues of the kernel defined by the finite network after 200 training epochs, as we progress through ResNet blocks.
    \label{fig:cifar10}
  }
\end{figure*}
The above results suggest that finite neural networks perform well by giving flexibility to interpolate between the input kernel and output kernel.
%To test the above theory, we extensively compared the kernels defined by infinite and finite variants of the same network architecture.
To see how this happens in real neural networks, we considered a 34-layer ResNet without batchnorm corresponding to the infinite network in \citet{garriga2019deep} trained on CIFAR-10.
We began by computing the correlation between elements of the finite and infinite kernel (Fig.~\ref{fig:cifar10}A top) as we go through ResNet blocks (x-axis), and as we go through training (blue lines).
As expected, the randomly initialized, untrained network retains a high correlation with the infinite kernel at all layers, though the correlation is somewhat smaller for higher layers, as there have been more time for random sampling to build up discrepancies.
However, for trained networks, this correspondence between the finite and infinite networks is far weaker: even at the first layer, the correlation is only around $0.5$, and as we go through layers, the correlation decreases to almost zero.
To understand whether kernels were being actively shaped, we computed the correlation between the kernel for the finite network and the output kernel, defined by taking the inner product of vectors representing the one-hot class labels (Fig.~\ref{fig:cifar10}A bottom).
We found that while the correlation for the untrained network decreased across layers, training gives strong positive correlations with the output kernel, and these correlations increase as we move through network layers.
Combined, these results indicate that the top-layer representation is much closer to the output kernel, as suggested by the deep linear results, than it is to the corresponding infinite network.
While correlation is a useful simple measure of similarity, there are other measures of similarity that take into account the special structure of kernel matricies.
In particular, we considered the marginal likelihood for the one-hot outputs corresponding to the class label, under a GP, with a kernel given by a scaled sum of the kernel at that ResNet block, and the identity (see Appendix~\ref{sec:app:natural-gradients}; Fig.~\ref{fig:cifar10}B top).
%To compute this measure, we maximized the marginal likelihood using a covariance that is a scaled sum of the kernel defined by that layer of the network and the identity 
For the infinite network, the marginal likelihood increased somewhat as we moved through network layers, and the untrained finite network performed similarly, except that there was a decrease in performance at the last layer. % up to the last layer, where there was a sharp drop-off in performance.
In contrast, the marginal likelihood for the finite, trained networks was initially very close to the infinite networks, but grew rapidly as we move through ResNet blocks.

To gain an insight into how training shaped the neural network kernels, we computed the variance in the subspace defined by the one-hot outputs (i.e. the classification directions; Fig.~\ref{fig:cifar10}B bottom).
We might have expected to see a steady increase in the variance in this subspace as we move through layers, but in fact the level was very small, only rising appreciably at the final block, and only for trained networks.
To try to understand these results, we computed the eigenvalue spectrum of the kernels.
For the infinite network (Fig.~\ref{fig:cifar10}C top), we found that the eigenvalue spectrum at all levels decayed as a $-1$ power law.
This is expected at the lowest level due to the well known $1/f$ power spectrum of images \citep{van1996modelling}, but is not necessarily the case at higher-levels.
Given the power-spectrum of the output kernel is just a small set of equal-sized eigenvalues corresponding to the class labels (Fig.~\ref{fig:cifar10}C bottom, green line), we might expect the eigenspectrum of finite networks to gradually get steeper as we move through network layers.
In fact, we find the opposite: for intermediate layers, the eigenvalue spectrum becomes flatter, which can be interpreted as the network attempting to retain as much information as possible about all aspects of the image.
It is only at the last layer where the relevant information is selected, giving an eigenvalue spectrum with around $10-1$ large and roughly equally-sized eigenvalues, followed by much smaller eigenvalues, which mirrors the spectrum of the output kernel.
This again confirms that the top-layer representation in trained networks is much closer to the output kernel than it is to corresponding infinite network.

\section{Related work}
\citet{agrawal2020wide} independently introduced infinite networks with finite bottlenecks, but then made a very different contribution in that context.
In particular, they highlighted that if we take the limit as some layers of a neural network go to infinity, convergence to the infinite networks with bottlenecks considered here is not immediate, but requires the neural network components to exhibit sufficient uniformity with respect to their inputs.
In contrast, we show that finite bottlenecks can introduce flexibility and thereby improve performance even in two-layer linear networks, give analytic results in the case of linear networks, and show that these considerations are likely to be important in realistic large-scale networks, by showing that the kernel for a trained ResNet differs dramatically from that for the corresponding infinite network.

Technically, our work bears similarity to classical work on the dynamics of gradient descent in unregularised deep linear networks \citep{saxe2013exact}.
Importantly, the lack of regularisation in this work implies that infinitely many optimal solutions are available (e.g.\ all the lower-layer weights being fixed to the identity).
In contrast, we focused on Bayesian inference, but also considered the optimal solution for regularised networks, which are much more constrained.

\section{Conclusions}
We have shown that finite Bayesian neural networks have more flexibility than infinite networks, and that this may explain the superior performance of finite networks.
Thus, we introduced infinite networks with bottlenecks, and argue that they may be as incorporate flexibility and are able to perform representation learning, they may be a better model of real neural networks.
We then assessed the flexibility of deep linear networks from two perspectives.
First, we looked at the prior viewpoint: the variability in the top-layer kernel induced by the prior over a finite neural network.
Second, we looked at the posterior viewpoint: the ability of the learning process to shape the top-layer kernel.
Under both MAP inference and sampling in finite networks, learning gradually shaped top-layer representations so as to match the output-kernel.
But, as Bayesian neural networks increase in width, the kernels become gradually less flexible, eliminating the possibility for learning to shape the kernel.
In contrast, for MAP inference, the degree of kernel shaping is not affected by network width, and this additional flexibility might be an avenue for overfitting.

\section*{Acknowledgements}
I would like to thank Adrià Garriga-Alonso, Sebastian Ober and Vidhi Lalchand for useful discussions.

\bibliography{refs}
\bibliographystyle{icml2020}

\appendix
\onecolumn
\section{Kernel flexibility: prior viewpoint}

To compute the covariance, which we denote $\Cov\sqb{\cdot}$ of the kernel for a deep network, we consider a recursion where we start with $\Cov\sqb{L^{\ell-1}_{ij}, L^{\ell-1}_{kl}| \L^0}$, then compute the resulting $\Cov\sqb{J^{\ell}_{ij}, J^{\ell}_{kl}| \L^0}$, then compute the resulting $\Cov\sqb{K^{\ell}_{ij}, K^{\ell}_{kl}| \L^0}$.
In particular, we apply the law of total covariance for $\K_\ell| \J_\ell$, and we consider linear networks for which $\L_\ell = \K_\ell$, %and fully connected networks for which $\Cov\sqb{J^{\ell}_{ij}, J^{\ell}_{kl}| \L^0} = \Cov\sqb{K^{\ell-1}_{ij}, K^{\ell-1}_{kl}| \L^0}$
%$\J_\ell| \L_{\ell-1}$, $\K_\ell| \J_\ell$ and $\L_\ell| \K_\ell$,
\begin{subequations}
\label{eq:rec:placeholder}
\begin{align}
  \label{eq:2:CovJ}
  \Cov\sqb{J^{\ell}_{ij}, J^{\ell}_{kl}| \L^0} &= \text{?}\\%\Cov\sqb{K^{\ell-1}_{ij}, K^{\ell-1}_{kl}| \L^0}\\
    %\Cov\sqb{\E\sqb{J^{\ell}_{ij}| \L^{\ell-1}}, \E\sqb{J^{\ell}_{kl}| \L^{\ell-1}}| \L^0}\\
  \label{eq:1:CovK}
  \Cov\sqb{K^{\ell}_{ij}, K^{\ell}_{kl}| \L^0} &= 
    \Cov\sqb{\E\sqb{K^{\ell}_{ij}| \J^{\ell}}, \E\sqb{K^{\ell}_{kl}| \J^{\ell}}| \L^0} 
    +\E\sqb{\Cov\sqb{K^{\ell}_{ij}, K^{\ell}_{kl}| \J^{\ell}}| \L^0}\\
  \Cov\sqb{L^{\ell}_{ij}, L^{\ell}_{kl}| \L^0} &= 
    \Cov\sqb{K^{\ell}_{ij}, K^{\ell}_{kl}| \L^0}
    %\Cov\sqb{\E\sqb{L^{\ell}_{ij}| \K^{\ell}}, \E\sqb{L^{\ell}_{kl}| \K^{\ell}}| \L^0}
\end{align}
\end{subequations}
The first equation is different for fully connected and convolutional networks, so we give its form later.

Eq.~\eqref{eq:1:CovK} always behaves in the same way for linear and nonlinear, fully connected and convolutional networks so we consider this first.
In particular, we always have $\E\sqb{\K_\ell| \J_\ell} = \J_\ell$, so the first term in Eq.~\eqref{eq:1:CovK} is
\begin{align}
  \label{eq:1:CovK:term1}
  \Cov\sqb{\E\sqb{K^{\ell}_{ij}| \J^{\ell}}, \E\sqb{K^{\ell}_{kl}| \J^{\ell}}| \L^0} &= \Cov\sqb{J^{\ell}_{ij}, J^{\ell}_{kl}| \L^0}.
\end{align}
For the second term in Eq.~\eqref{eq:1:CovK}, we substitute the definition of $\K_\ell$ (Eq.~\ref{eq:def:KL}) into the definition of the covariance,
\begin{multline}
  \Cov\sqb{K^{\ell}_{ij}, K^{\ell}_{kl}| \J^{\ell}} = \tfrac{1}{N_\ell^2} \sum_{\mu=1}^{N_\ell} \sum_{\nu=1}^{N_\ell} \E\sqb{ a^\ell_{\mu i} a^\ell_{\mu j} a^\ell_{\nu k} a^\ell_{\nu l}| \J^{\ell}}\\ 
  - \b{\tfrac{1}{N_\ell} \sum_{\mu=1}^{N_\ell} \E\sqb{a^\ell_{\mu i} a^\ell_{\mu j}| \J^{\ell}}} \b{ \tfrac{1}{N_\ell} \sum_{\nu=1}^{N_\ell} \E\sqb{a^\ell_{\nu k} a^\ell_{\nu l}| \J^{\ell}}}.
\end{multline}
As the $a$'s are jointly Gaussian, their expectations are
\begin{subequations}
\begin{align}
  \E\sqb{a^{\ell}_{\mu i} a^{\ell}_{\mu j} a^{\ell}_{\nu k} a^{\ell}_{\nu l}| \J^\ell} &= J^\ell_{ij} J^\ell_{kl} + \delta_{\mu\nu} \b{J^\ell_{ik} J^\ell_{jl} + J^\ell_{il} J^\ell_{jk}}\\
  \E\sqb{a^{\ell}_{\mu i} a^{\ell}_{\mu j}| \J^\ell} &= J^\ell_{ij} \\ 
  \E\sqb{a^{\ell}_{\nu k} a^{\ell}_{\nu l}| \J^\ell} &= J^\ell_{kl}.
\end{align}
\end{subequations}
Thus, the covariance of the kernel becomes,
\begin{align}
  \label{eq:Cov:K|J}
  %\Cov\sqb{K^{\ell}_{ij}, K^{\ell}_{kl}| \K_{\ell-1}} = \Cov\sqb{K^{\ell}_{ij}, K^{\ell}_{kl}| \J_{\ell}} 
  \Cov\sqb{K^{\ell}_{ij}, K^{\ell}_{kl}| \J_{\ell}} &= \tfrac{1}{N_\ell} \b{J^\ell_{ik} J^\ell_{jl} + J^\ell_{il} J^\ell_{jk}},
\end{align}
Substituting this into the second term in Eq.~\eqref{eq:1:CovK}
%Substituting this into the second term in Eq.~\eqref{eq:1:CovK} and writing the expected product in terms of the product of expectations,
\begin{align}
  \E\sqb{\Cov\sqb{K^{\ell}_{ij}, K^{\ell}_{kl}| \J_{\ell}}| \L_0} &= \tfrac{1}{N_\ell} \E\sqb{J^\ell_{ik} J^\ell_{jl} + J^\ell_{il} J^\ell_{jk}| \L_0},
\end{align}
and writing the expected product in terms of the product of expectations and covariance,
\begin{align}
  \label{eq:1:CovK:term2}
  \E\sqb{\Cov\sqb{K^{\ell}_{ij}, K^{\ell}_{kl}| \J_{\ell}}| \L_0} &=\tfrac{1}{N_\ell} \b{\ab{J^\ell_{ik}}  \ab{J^\ell_{jl}} + \ab{J^\ell_{il}} \ab{J^\ell_{jk}}}
  + \tfrac{1}{N_\ell} \b{ \Cov\sqb{J^\ell_{ik}, J^\ell_{jl}| \L_0} + \Cov\sqb{J^\ell_{il}, J^\ell_{jk}| \L_0}}
  \intertext{where,}
  \ab{J^\ell_{ik}} &= \E\sqb{J^\ell_{ik}| \L_0}.
\end{align}
Substituting Eq.~\eqref{eq:1:CovK:term1} and Eq.~\eqref{eq:1:CovK:term2} into Eq.~\eqref{eq:1:CovK}, we obtain
\begin{multline}
  \label{eq:2:CovK}
  \Cov\sqb{K^{\ell}_{ij}, K^{\ell}_{kl}| \L^0} = \Cov\sqb{J^{\ell}_{ij}, J^{\ell}_{kl}| \L^0}
  + \tfrac{1}{N_\ell} \b{\ab{J^\ell_{ik}}  \ab{J^\ell_{jl}} + \ab{J^\ell_{il}} \ab{J^\ell_{jk}}}\\
  + \tfrac{1}{N_\ell} \b{ \Cov\sqb{J^\ell_{ik}, J^\ell_{jl}| \L_0} + \Cov\sqb{J^\ell_{il}, J^\ell_{jk}| \L_0}}
\end{multline}
%We could directly use this expression but it is somewhat unwieldy as the right-hand-side contains three covariance terms.
%
%
%As such, note that it is possible to approximate this expression as the final two covariance terms are $\mathcal{O}(1/N^2)$, in comparison to $\mathcal{O}(1/N)$ for the other terms, and as such they become negligable as $N$ grows,
%\begin{multline}
%  \label{eq:approx:CovK}
%  \Cov\sqb{K^{\ell}_{ij}, K^{\ell}_{kl}| \L^0} \approx \Cov\sqb{J^{\ell}_{ij}, J^{\ell}_{kl}| \L^0}
%  + \tfrac{1}{N_\ell} \b{\ab{J^\ell_{ik}}  \ab{J^\ell_{jl}} + \ab{J^\ell_{il}} \ab{J^\ell_{jk}}}.\hfill
%\end{multline}
%Using Eq.~\eqref{eq:approx:CovK} in place of Eq.~\eqref{}, and using 
%
%for a linear network, we obtain the expressions used in the main text.
%
%\begin{subequations}
%  \label{eq:cov-rec}
%\begin{align}
%  \label{eq:2:CovJ}
%  \Cov\sqb{J^{\ell}_{ij}, J^{\ell}_{kl}| \L^0} &= \text{?}\\
%  \Cov\sqb{K^{\ell}_{is,jr}, K^{\ell}_{kl}| \L^0} &\approx
%    \Cov\sqb{J^{\ell}_{ij}, J^{\ell}_{kl}| \L^0} 
%    +\tfrac{1}{N_\ell} \b{\ab{J^\ell_{ik}} \ab{J^\ell_{jl}} + \ab{J^\ell_{il}} \ab{J^\ell_{jk}}}\\
%  \Cov\sqb{L^{\ell}_{ij}, L^{\ell}_{kl}| \L^0} &= 
%    \Cov\sqb{K^{\ell}_{ij}, K^{\ell}_{kl}| \L^0}
%\end{align}
%\end{subequations}

\subsection{Fully connected network}
Now we evaluate Eq.~\eqref{eq:2:CovJ} first for a fully connected network,
%For a fully connected, network,
\begin{align}
  a^{\ell}_{\lambda, i} &= \sum_{\mu} h^{\ell-1}_{i, \mu} W^{\ell}_{\mu, \lambda} 
  \intertext{where the weights are drawn from an independent zero-mean Gaussian, such that}
  \label{eq:covW:fc}
  E\sqb{W^{\ell}_{\mu, \lambda} W^{\ell}_{\nu, \lambda}} &= \tfrac{1}{N_{\ell-1}} \delta_{\mu, \nu}
  \intertext{Thus, $\a^{\ell}_\lambda$ has distribution,}
  \P{\a^{\ell}_\lambda} &= \N{\a^{\ell}_\lambda; \0, \J^{\ell}},
  \intertext{where $\J^{\ell}$ is given by,}
  J^{\ell}_{ij} &= \Cov\sqb{a^{\ell}_{i, \lambda}, a^{\ell}_{j, \lambda}} 
                    = \E\sqb{a^{\ell}_{i, \lambda} a^{\ell}_{j, \lambda}}\\
  &= \E\sqb{\b{\sum_{\mu} h^{\ell-1}_{i, \mu} W^{\ell}_{\mu, \lambda}} \b{\sum_{\nu} h^{\ell-1}_{j, \nu} W^{\ell}_{\nu, \lambda}}}\\
  &= \sum_{\mu \nu} h^{\ell-1}_{i, \mu} h^{\ell-1}_{j, \nu} \E\sqb{W^{\ell}_{\mu, \lambda} W^{\ell}_{\nu, \lambda}} 
  \intertext{substituting for the expectation (Eq.~\ref{eq:covW:fc}), and identifying the activity kernel (Eq.~\ref{eq:def:KL}),}
  &= \tfrac{1}{N_{\ell-1}} \sum_{\mu} h^{\ell-1}_{i, \mu} h^{\ell-1}_{j, \mu} = L^{\ell-1}_{ij} %= K^{\ell-1}_{ij}
\end{align}
Thus,
\begin{align}
  \Cov\sqb{J^{\ell}_{ij}, J^{\ell}_{kl}| \K^0} &= \Cov\sqb{L^{\ell-1}_{ij}, L^{\ell-1}_{kl} |\K^0}
\end{align}
Combining this expression with Eq.~\eqref{eq:2:CovK} gives a complete form for the updates (Eq.~\ref{eq:rec:placeholder}),
\begin{subequations}
\label{eq:rec:fc-complex}
\begin{align}
  \Cov\sqb{J^{\ell}_{ij}, J^{\ell}_{kl}| \L^0} =& \Cov\sqb{L^{\ell-1}_{ij}, L^{\ell-1}_{kl} |\K^0}\\
  \nonumber
  \label{eq:3:CovK}
  \Cov\sqb{K^{\ell}_{is,jr}, K^{\ell}_{kl}| \L^0} =&
    \Cov\sqb{J^{\ell}_{ij}, J^{\ell}_{kl}| \L^0} 
    +\tfrac{1}{N_\ell} \b{\ab{J^\ell_{ik}} \ab{J^\ell_{jl}} + \ab{J^\ell_{il}} \ab{J^\ell_{jk}}}\\
    &+ \tfrac{1}{N_\ell} \b{ \Cov\sqb{J^\ell_{ik}, J^\ell_{jl}| \L_0} + \Cov\sqb{J^\ell_{il}, J^\ell_{jk}| \L_0}}\\
  \Cov\sqb{L^{\ell}_{ij}, L^{\ell}_{kl}| \L^0} =& 
    \Cov\sqb{K^{\ell}_{ij}, K^{\ell}_{kl}| \L^0}
\end{align}
\end{subequations}

However, this form is difficult to analyse due to the complexity of Eq.~\eqref{eq:3:CovK}.
Instead we can form an approximation to Eq.~\eqref{eq:3:CovK} by noting that one of the recursive terms is negligable.
Taking all network widths to be equal, $N_\ell = N$, (or at least of the same order), if
%In particular, note that for $\ell=1$, the covariance terms are zero as the data is fixed, so,
%\begin{align}
%  \Cov\sqb{J^{1}_{ij}, J^{1}_{kl}| \L^0} &= \Cov\sqb{L^{0}_{ij}, L^{0}_{kl}| \L^0} = 0\\
%  \intertext{thus, there is one only non-zero term in Eq.~\ref{} for $\ell=1$, and this term is $\mathcal{O}(1/N_\ell)$, as the network is chosen such that the activities, and hence the covariances, $J^\ell_{ij}$ remain $\mathcal{O}(1)$,}
%  \Cov\sqb{K^{1}_{ij}, K^{1}_{kl}| \L^0} &= \tfrac{1}{N} \b{\ab{L^0_{ik}}  \ab{L^0_{jl}} + \ab{J^0_{il}} \ab{J^0_{jk}}} = \mathcal{O}(1/N).
%\end{align}
%recursively applying Eq.~\eqref{eq:rec:fc-complex}, we see that if 
\begin{align}
  \Cov\sqb{L^{\ell-1}_{ij}, L^{\ell-1}_{kl}| \L^0} &= \mathcal{O}(1/N),
  \intertext{then}
  \Cov\sqb{J^{\ell}_{ij}, J^{\ell}_{kl}| \L^0} &= \mathcal{O}(1/N),
  \intertext{as the network is chosen such that the activities, and hence the covariances, $J^\ell_{ij}$ remain $\mathcal{O}(1)$}
  \label{eq:approx:CovK}
  \Cov\sqb{K^{\ell}_{is,jr}, K^{\ell}_{kl}| \L^0} =&
    \Cov\sqb{J^{\ell}_{ij}, J^{\ell}_{kl}| \L^0} 
    +\tfrac{1}{N_\ell} \b{\ab{J^\ell_{ik}} \ab{J^\ell_{jl}} + \ab{J^\ell_{il}} \ab{J^\ell_{jk}}} + \mathcal{O}(1/N^2) = \mathcal{O}(1/N),
  \intertext{so,}
  \Cov\sqb{L^{\ell}_{ij}, L^{\ell}_{kl}| \L^0} &= \mathcal{O}(1/N).
\end{align}
To begin the recursion, the data is fixed, so
%In particular, note that for $\ell=1$, the covariance terms are zero as the data is fixed, so,
\begin{align}
  \Cov\sqb{J^{1}_{ij}, J^{1}_{kl}| \L^0} &= \Cov\sqb{L^{0}_{ij}, L^{0}_{kl}| \L^0} = 0,
  \intertext{and,}
  \Cov\sqb{K^{1}_{ij}, K^{1}_{kl}| \L^0} &= \tfrac{1}{N} \b{\ab{L^0_{ik}}  \ab{L^0_{jl}} + \ab{J^0_{il}} \ab{J^0_{jk}}} = \mathcal{O}(1/N).
\end{align}
Thus, the covariance of the kernels and covariances is in indeed $\mathcal{O}(1/N)$, so $\Cov{K^\ell_{ij}, K^\ell_{kl}}$ can be approximated by Eq.~\eqref{eq:approx:CovK}. 
Combining this approximation with Eq.~\eqref{eq:rec:fc-complex} gives the expressions in the main text (Eq.~\ref{eq:prior}).
Finally, note that the approximation in Eq.~\eqref{eq:approx:CovK} remains true only as long as the number of layers is small, $L \ll N$.
%Using Eq.~\eqref{eq:approx:CovK} instead of Eq.~\eqref{eq:3:CovK} gives the expressions in the main text (Eq.~\ref{eq:prior}).

%which can be substituted into the recursion Eq.~\eqref{eq:2:CovJ} allowing the updates to be directly computed,
%\begin{subequations}
%\label{eq:prior:fc}
%\begin{align}
%  \Cov\sqb{J^{\ell}_{ij}, J^{\ell}_{kl}| \L^0} &= \Cov\sqb{L^{\ell-1}_{ij}, L^{\ell-1}_{kl} |\K^0}\\
%  \Cov\sqb{K^{\ell}_{is,jr}, K^{\ell}_{kl}| \L^0} &\approx
%    \Cov\sqb{J^{\ell}_{ij}, J^{\ell}_{kl}| \L^0} 
%    +\tfrac{1}{N_\ell} \b{\ab{J^\ell_{ik}} \ab{J^\ell_{jl}} + \ab{J^\ell_{il}} \ab{J^\ell_{jk}}}\\
%  \Cov\sqb{L^{\ell}_{ij}, L^{\ell}_{kl}| \L^0} &= 
%    \Cov\sqb{K^{\ell}_{ij}, K^{\ell}_{kl}| \L^0}
%\end{align}
%\end{subequations}

\subsection{Convolutional network}
\label{app:prior:conv}  
For locally connected and convolutional networks, we introduce spatial structure into the activations, and we use spatial indicies, $r$, $s$, $u$ and $v$.
Thus, the activations for datapoint $i$ at layer $\ell$, spatial location $r$ and channel $\lambda$ are given by,
\begin{align}
  a^{\ell}_{i,r\lambda} &= \sum_{r'\mu} h^{\ell-1}_{i,r'\mu} W^{\ell}_{r' \mu, r\lambda}.
\end{align}
Note that for many purposes, these higher-order tensors can be treated as vectors and matrices, if we combine indicies (e.g. using a ``reshape'' or ``view'' operation).  The commas in the index list are used to denote how to combine indicies for this particular operation, such that it can be understood as a standard matrix/vector operation.
For the above equation, the activations, $\a^\ell \in \mathbb{R}^{P \times SN_\ell}$ are given by the matrix product of the activities from the previous layer, $\h^{\ell-1} \in \mathbb{R}^{P\times SM_{\ell-1}}$, and the weights, $\W_\ell \in \mathbb{R}\in^{SM_{\ell-1} \times SN_\ell}$, where remember that $S$ is the number of spatial locations in the input.

For a convolutional neural network, the weights are the same if we consider the same input-to-output channels, and the same spatial displacement, $d$, and are uncorrelated otherwise,
\begin{align}
  %E\sqb{W^{\ell}_{r' \mu, \lambda r} W^{\ell}_{\nu s', \lambda s}} &= \tfrac{1}{N_{\ell-1} D_{\ell-1}} \delta_{\mu, \nu} \sum_{d\in\mathcal{D}_{\ell-1}} \delta_{d,(r-r')} \delta_{d,(s-s')}.
  \label{eq:E[WWT]:cnn}
  E\sqb{W^{\ell}_{r' \mu, r \lambda} W^{\ell}_{s' \nu, s \lambda}} &= \tfrac{1}{M_{\ell-1} D_{\ell-1}} \delta_{\mu, \nu} \sum_{d\in\mathcal{D}_{\ell-1}} \delta_{r',(r+d)} \delta_{s',(s+d)}.
\end{align}
where $\mathcal{D}_{\ell-1}$ is the set of all valid spatial displacements for the convolution, and $D_{\ell-1} = \abs{\mathcal{D}_{\ell-1}}$ is the number of valid spatial displacements (i.e.\ the size of the convolutional patch).
For a locally-connected network, the only additional requirement is that the output spatial locations are the same,
\begin{align}
  \label{eq:E[WWT]:lcn}
  %E\sqb{W^{\ell}_{r' \mu, \lambda r} W^{\ell}_{\nu s', \lambda s}} &= \tfrac{1}{N_{\ell-1} D_{\ell-1}} \delta_{\mu, \nu} \delta_{r, s} \sum_{d\in\mathcal{D}_{\ell-1}} \delta_{d,(r-r')} \delta_{d,(s-s')}.
  E\sqb{W^{\ell}_{r' \mu, r \lambda} W^{\ell}_{s' \nu, s \lambda}} &= \tfrac{1}{M_{\ell-1} D_{\ell-1}} \delta_{\mu, \nu} \delta_{r, s} \sum_{d\in\mathcal{D}_{\ell-1}} \delta_{r',(r+d)} \delta_{s',(s+d)}.
\end{align}

Now we can compute the covariance of the activations, $\J^\ell$, for a convolutional network,
\begin{align}
  %\P{\a^{\ell}_\lambda} &= \N{\a^{\ell}_\lambda; \0, \J^{\ell}},
  %\intertext{where $\J^{\ell}$ is given by,}
  J^{\ell}_{ir, js} &= \E\sqb{a^{\ell}_{i,r\lambda} a^{\ell}_{j,s\lambda}| \L_{\ell-1}}\\
  &= \E\sqb{\left. \b{\sum_{r' \mu} h^{\ell-1}_{i, r' \mu} W^{\ell}_{r'\mu, r\lambda}} \b{\sum_{s'\nu} h^{\ell-1}_{j, s' \nu} W^{\ell}_{s' \nu, r \lambda}} \right| \L_{\ell-1}}\\
  &= \sum_{\mu\nu r's'}h^{\ell-1}_{i, r'\mu} h^{\ell-1}_{j, s'\nu} \E\sqb{W^{\ell}_{r'\mu, r\lambda} W^{\ell}_{s' \nu, r \lambda}} \\
  \intertext{substituting the covariance of the weights (Eq.~\ref{eq:E[WWT]:cnn}), and noting that the product of $h$'s forms the definition of the activity kernel (Eq.~\ref{eq:def:KL}),}
  \label{eq:cnn:conv:J|K}
  J^{\ell}_{ir, js} &= \tfrac{1}{D_{\ell-1}} \sum_{d\in\mathcal{D}_{\ell-1}} L^{\ell-1}_{i(r+d), j(s+d)}
\end{align}
%where the last equality is given by substituting the  and that, for a linear network, $\K_\ell = \L_\ell$.
For locally connected intermediate layers, we instead substitute Eq.~\eqref{eq:E[WWT]:lcn}, which gives the same result, except that the output locations must be the same for there to be any covariance in the weights,
\begin{align}
  \label{eq:lcn:conv:J|K}
  J^{\ell}_{ir, js} = \tfrac{1}{D_{\ell-1}} \delta_{r, s} \sum_{d\in\mathcal{D}_{\ell-1}} L^{\ell-1}_{i(r+d), j(s+d)}
\end{align}
Substituting this into Eq.~\eqref{eq:2:CovJ},
\begin{align}
  \Cov\sqb{J^{\ell}_{ir,js}, J^{\ell}_{ku,lv}| \L^0} = 
  \Cov\sqb{
    \tfrac{1}{D_{\ell-1}} \sum_{d\in\mathcal{D}_{\ell-1}} L^{\ell-1}_{i(r+d), j(s+d)},
    \tfrac{1}{D_{\ell-1}} \sum_{d\in\mathcal{D}_{\ell-1}} L^{\ell-1}_{k(u+d), l(v+d)}
  }.
\end{align}
Now, we can put together full recursive updates for convolutional networks, by pulling the sum out of the covariance above, and by taking the indicies in Eq.~\eqref{eq:rec:placeholder}, as indexing both a datapoint and a spatial location (i.e.\ $i \rightarrow i,s$),
\begin{subequations}
\label{eq:prior:conv}
\begin{align}
  \Cov\sqb{J^{\ell}_{ir,js}, J^{\ell}_{ku,lv}| \L^0} &= 
  \tfrac{1}{D_{\ell-1}^2} \sum_{dd'} \Cov\sqb{L^{\ell-1}_{i(r+d),j(s+d)}, L^{\ell-1}_{k(u+d'),l(v+d')}| \L^0}\\
  \Cov\sqb{K^{\ell}_{ir,js}, K^{\ell}_{ku,lv}| \L^0} &\approx
    \Cov\sqb{J^{\ell}_{ir,js}, J^{\ell}_{ku,lv}| \L^0} 
    +\tfrac{1}{N_\ell} \b{\ab{J^\ell_{ir,ku}} \ab{J^\ell_{js,lv}} + \ab{J^\ell_{ir,lv}} \ab{J^\ell_{js,ku}}}\\
  \Cov\sqb{L^{\ell}_{ir,js}, L^{\ell}_{ku,lv}| \L^0} &= 
    \Cov\sqb{K^{\ell}_{ir,js}, K^{\ell}_{ku,lv}| \L^0}
\end{align}
\end{subequations}
Finally, to compute these terms, note that we can recursively compute these expressions for $r=s$ and $u=v$, 
\begin{align}
  \label{eq:CovJ:space}
  \Cov\sqb{J^{\ell}_{ir,jr}, J^{\ell}_{ku,lu}| \L^0} &= 
  \tfrac{1}{D_{\ell-1}^2} \sum_{dd'} \Cov\sqb{L^{\ell-1}_{i(r+d),j(r+d)}, L^{\ell-1}_{k(u+d'),l(u+d')}| \L^0},
\end{align}
which reduces computational complexity, and the resulting expression can even be evaluated efficiently as a 2D convolution.

\subsubsection{Convolutional and locally connected networks}
\label{sec:app:space}
To understand the very different results for convolutional and locally structured networks (Fig.~\ref{fig:prior}B--D) despite their having the same infinite limit, we need to consider how Eq.~\eqref{eq:CovJ:space} interacts with Eq.~\eqref{eq:lcn:conv:J|K}.
For a locally connected network, the covariance of activations at different locations is always zero, i.e.\ $J_{ir, js}^\ell = 0$ for $r\neq s$ whereas, for a spatially structured network, the $J_{ir, js}$ terms for $r\neq s$ have the same scale as those for $r=s$.
The $J_{ir, js}$ terms enter into the variance of the kernel through Eq.~\eqref{eq:CovJ:space}.  
Note that there are $D_{\ell-1}^2$ terms in this sum, and the sum is normalized by dividing by $D_{\ell-1}^2$.
Thus, in convolutional networks, there are $D_{\ell-1}^2$ terms all with the same scale, whereas in spatially unstructured networks, we have only $D_{\ell-1}$ nonzero terms, introducing an effective $1/D_{\ell-1}$ normalizer.
%so $\Cov[J^{\ell}_{ir,jr}, J^{\ell}_{ku,lu}| \L^0]$ is $\mathcal{O}(1/D_{\ell-1})$.
This is particularly important if we consider the last layer.
The last layer can be understood as a convolution, where the convolutional patch has the same size as the image (i.e.\ $D_L = S$), and there is no padding, such the the output has a single spatial location.
In this case, the $1/S = 1/D_L$ normalizer can be very large. %, which causes the second term in Eq.~\eqref{eq:CovK} to dominate the first term, leading to the almost constant variance of the kernel as we increase network depth observed in Fig.~\ref{fig:prior}D.

\section{Kernel flexibility: posterior viewpoint}
\subsection{Reparameterising finite neural networks}
Swapping between a kernel representation and a feature representation is difficult if we work directly with a prior over the weights, $\W_\ell \in \mathbb{R}^{N_{\ell-1} \times N_{\ell}}$.
Instead, note that as the weights are Gaussian, we can reparameterise the neural network, working instead with $\V_\ell \in \mathbb{R}^{P \times N_\ell}$ which has independent standard Gaussian entries, where $P$ is the number of datapoints.
In particular, we can write the activities at the next layer using,
\begin{align}
  \label{eq:A=LV}
  \A_\ell &= \H_{\ell-1} \W_\ell = \U_{\ell}^T \V_\ell.
\end{align}
 where $\U_\ell \in \mathbb{R}^{P \times P}$ is any matrix that satisfies,
\begin{align}
  \J_\ell &= \U_\ell^T \U_\ell.
\end{align}
such as the Cholesky decomposition of the covariance, $\J_\ell$.
We can thus write the kernel as,
\begin{align}
  \label{eq:app:K_V}
  \K_{\ell} &= \tfrac{1}{N_\ell} \A_{\ell} \A_{\ell}^T = \tfrac{1}{N_\ell} \H_{\ell-1} \W_\ell \W_\ell^T \H_{\ell-1}^T = \tfrac{1}{N_\ell} \U_{\ell}^T \V_\ell \V_\ell^T \U_{\ell}.
\end{align}
Rearranging, we can write $\tfrac{1}{N_\ell} \V_\ell \V_\ell^T$, or equivalently the mismatch between the covariance, $\J_\ell$, and the output kernel, $\K_\ell$, in terms of $\L_\ell$ and $\K_\ell$, and we denote this quantity $\R_\ell$ for future use,
\begin{align}
  \R_\ell = \U_{\ell}^{-T} \K_{\ell} \U_{\ell}^{-1} &= \tfrac{1}{N_\ell} \V_\ell \V_\ell^T 
\end{align}
where $\X^{-T} = (\X^{-1})^T = (\X^T)^{-1}$, and we have assumed that $\J_\ell$ is invertible, which if nothing else, requires that the number of features $M_\ell$ and $N_\ell$ are larger than the number of datapoints.

%\section{When is it valid to throw away rotations?}
%The above derivation gives us a way to compute the activation kernel, $\K_\ell$, directly from the covariance, $\J_\ell$, without needing to use the activations themselves.
%To work with the whole network in kernel space, we need to be able to compute the activity kernel, $\L_\ell$, from the activation kernel, $\K_\ell$.
%However, this is not possible in general, because computing the kernel throws away the orientation in activity space, and this matters for standard neural network activation functions.
%That said, there are two circumstances where it is possible to compute $\L_\ell$ from $\K_\ell$.
%First, in a linear network,
%\begin{align}
%  \L_\ell &= \K_\ell,
%\end{align}
%and while this network is not practically useful, it can be used to give strong intuitions about how nonlinear networks function.
%Alternatively, we can consider a finite-infinite network, which involves 

\subsection{MAP inference}
\label{app:post:map}

Here, we consider MAP inference over $\V_\ell$.
As the entries of $\V_\ell$ have a standard Gaussian prior, we have,
\begin{align}
  \log \P{\V_\ell} &= -\tfrac{1}{2} \tr\b{\V_\ell \V_\ell^T} + \const \\
  &= -\tfrac{N_\ell}{2}\tr\b{\U^{-T}_{\ell} \K_{\ell} \U_{\ell}^{-1}} + \const \\
  &= -\tfrac{N_\ell}{2}\tr\b{\K_{\ell} \U_{\ell}^{-1}\U^{-T}_{\ell}} + \const \\
  &= -\tfrac{N_\ell}{2}\tr\b{\K_{\ell} \b{\U_{\ell}^T\U_{\ell}}^{-1}} + \const \\
  &= -\tfrac{N_\ell}{2}\tr\b{\K_{\ell} \J_{\ell}^{-1}} + \const
\end{align}

We can write the likelihood in the same form,
\begin{align}
  \log \P{\Y| \J_{L+1}} &= -\tfrac{1}{2} \tr\b{\Y^T \J_{L+1}^{-1} \Y} + \const\\
  \log \P{\Y| \J_{L+1}} &= -\tfrac{1}{2} \tr\b{\Y \Y^T \J_{L+1}^{-1}} + \const\\
  \log \P{\Y| \J_{L+1}} &= -\tfrac{N_{L+1}}{2} \tr\b{\K_{L+1} \J_{L+1}^{-1}} + \const\\
  \intertext{where,}
  \K_{L+1} &= \tfrac{1}{N_{L+1}} \Y \Y^T.
\end{align}
Note that we would usually incorporate IID noise in the outputs, and we are not doing so here in order to give exact, interpretable solutions.
We do not expect this to change the overall pattern of the results, except to marginally weaken the connection between the the output-kernel, $\K_{L+1}$, and top-layer kernel, $\K_L$.

Thus, the joint probability can be written as,
\begin{align}
  \log \P{\V_1,\dotsc,\V_L,\Y| \X} &= -\tfrac{1}{2} \sum_{\ell=1}^{L+1} N_\ell\tr\b{\K_{\ell} \J_{\ell}^{-1}} + \const.
\end{align}
Now we find, the MAP values of $\V_1,\dotsc\V_L$
\begin{align}
  \V^*_1,\dotsc,\V^*_L &= \argmax_{\V_1,\dotsc,\V_L} \log \P{\V_1,\dotsc,\V_L,\Y| \X},
\end{align}
by taking gradients of $\P{\V_1,\dotsc,\V_L,\Y| \X}$ wrt $\K_1,\dotsc,\K_L$.
Note that we can find the mode of this distribution by differeniating with respect to many different quantities, and we choose $\K_\ell$ because of algebraic convenience, and because it includes all relevant information from $\V_\ell$ (Eq.~\ref{eq:app:K_V}).
Further, note that as we are still working with the probability density of $\V_1,\dotsc,\V_L$ we should not include a Jacobian term.
Now we consider a linear, fully connected network where $\J_{\ell} = \K_{\ell-1}$, 
\begin{align}
  \0 = \dd{\K_\ell}\log \P{\V_1,\dotsc,\V_L,\Y| \X} &= - \tfrac{N_\ell}{2} \K_{\ell-1}^{-1} + \tfrac{N_{\ell+1}}{2} \K_\ell^{-1} \K_{\ell+1} \K_\ell^{-1}
  \intertext{where we have used,}
  \dd[\tr\b{\K_{\ell}^{-1} \K_{\ell+1}}]{\K_\ell} &= -\K_{\ell}^{-1} \K_{\ell+1} \K_{\ell}^{-1}\\
  \dd[\tr\b{\K_{\ell-1}^{-1} \K_\ell}]{\K_\ell} &= \K_{\ell-1}^{-1}
\end{align}
Thus, the MAP kernel changes as a fixed ratio,
\begin{align}
  \mS = N_{\ell+1} \K_{\ell+1} \K_{\ell}^{-1} &= N_\ell \K_{\ell} \K_{\ell-1}^{-1},
\end{align}
As the input kernel, $\K_0$, and the output kernel, $\K_{L+1}$, are fixed we can solve for $\mS$,
\begin{align}
  %\K_{L+1} \K_0^{-1} &= \prod_{\ell=1}^{L+1} \K_{\ell} \K_{\ell-1},\\
  %\intertext{multiplying by all $\N_\ell$ terms on both sides,}
  \mS^{L+1} &= \prod_{\ell=1}^{L+1} N_\ell \K_{\ell} \K_{\ell-1}^{-1} =\K_{L+1} \K_0^{-1} \prod_{\ell=1}^{L+1} N_\ell
  \intertext{so,}
  \mS &= \b{\K_{L+1} \K_0^{-1}}^{1/L+1} \b{\prod_{\ell=1}^{L+1} N_\ell}^{1/L+1}
\end{align}
where the final term is the geometric average of the width at each layer.
As such, the kernel at any given layer is,
\begin{align}
  \K_\ell &= \b{\prod_{\ell'=1}^\ell \K_\ell \K_{\ell-1}^{-1}} \K_0\\
  &= \b{\prod_{\ell'=1}^\ell \tfrac{1}{N_{\ell'}} \mS} \K_0\\
  &= \frac{\b{\prod_{\ell'=1}^{L+1} N_{\ell'}}^{\ell/(L+1)}}{\prod_{\ell'=1}^\ell N_{\ell'}} \b{\K_{L+1} \K_0^{-1}}^{\ell/(L+1)}  \K_0
  \intertext{defining the geometric average of the number of units at each layer prior to (and including) $\ell$, and after $\ell$,}
  N_{\leq \ell} &= \b{\prod_{\ell'=1}^{\ell} N_{\ell'}}^{1/\ell}\\
  N_{\ell <} &= \b{\prod_{\ell'=\ell+1}^{L+1} N_{\ell'}}^{1/(L + 1 - \ell)}\\
  \intertext{we can write,}
  \frac{\b{\prod_{\ell'=1}^{L+1} N_\ell}^{\ell/(L+1)}}{\prod_{\ell'=1}^\ell N_\ell}
  &= \frac{\b{\b{N_{\leq \ell}}^\ell \b{N_{\ell<}}^{L+1-\ell}}^{\ell/(L+1)}}{\b{N_{\leq \ell}}^\ell}\\
  &= \b{\b{N_{\leq \ell}}^{-(L+1-\ell)} \b{N_{\ell<}}^{L+1-\ell}}^{\ell/(L+1)}\\
  &= \b{\frac{N_{\ell<}}{N_{\leq \ell}}}^{\tfrac{\ell (L+1-\ell)}{L+1}}
\end{align}
This factor is the ratio of the geometric average of the widths for the previous and subsequent layers, to a power which depends on the distance to the end points (for $\ell=0$ or $\ell=L+1$ this power is 0),
\begin{align}
  \K_\ell &= \b{\frac{N_{\ell<}}{N_{\leq \ell}}}^{\tfrac{\ell (L+1-\ell)}{L+1}} \b{\K_{L+1} \K_0^{-1}}^{\ell/(L+1)}  \K_0
\end{align}
Thus, MAP does something sensible: no matter the network widths (and including as the network widths go to infinity), the representation interpolates smoothly between the input and output kernels.
However, the scale of these representations can shift in a strange, and potentially pathological fashion.
Remember that we normalized the weights, taking into account the width of each layer such that the representations maintained the same scale, irrespective of layer width.
However, under MAP inference, the network width controls the scale of the kernel, with larger kernels at layer $\ell$ given by widening layers from $1$ to $\ell$, and narrowing layers from $\ell+1$ to $L+1$.

\section{Deriving a cost-function such that gradient descent is equivalent to sampling}
\label{app:post:langevin}
The pathologies in the above derivations indicate that MAP, using full-batch gradient descent may give a very poor approximation of the kernel induced by \textit{stochastic} gradient descent.
As such, we consider Langevin sampling which not only gives Bayesian inference, but also gives a good starting point for thinking about the noise introduced by stochastic gradient descent.
In particular, we perform Langevin sampling over $\V_\ell$ (Eq.~\ref{eq:A=LV})
\begin{align}
  d \V_\ell &= \tfrac{1}{2} dt \dd[\mathcal{L}]{\V_\ell} + d\mathbf{\Xi}_\ell,
\end{align}
where $d\mathbf{\Xi}_\ell$ is a matrix-valued Weiner process.
Remembering that the objective is completely specified by $\R_\ell = \tfrac{1}{N_\ell} \V_\ell \V_\ell^T$, for a linear or finite-infinite network, we consider the effect of this sampling on $\R_\ell$.
In particular, we consider the expected change in $\R_\ell$ under Langevin sampling,
\begin{align}
  \label{eq:def:dR}
  \E\sqb{d \R_\ell| \R_\ell} &= 
  \E\sqb{\tfrac{1}{N_\ell} d\b{\V_\ell \V_\ell^T}} = 
  \tfrac{1}{2 N_\ell} dt \b{\dd[\mathcal{L}]{\V_\ell} \V_\ell^T + \V_\ell \dd[\mathcal{L}]{\V_\ell}^T} + \tfrac{1}{N_\ell} \E\sqb{d\mathbf{\Xi}_\ell d\mathbf{\Xi}_\ell^T}.
\end{align}
As the only stochasticity comes from the last term, and this term has known expectation,
\begin{align}
  \tfrac{1}{N_\ell} \E\sqb{d\mathbf{\Xi}_\ell d\mathbf{\Xi}_\ell^T} &= dt \; \I,
\end{align}
We can compute the expected update, which becomes the exact update as we take $N_\ell \rightarrow \infty$,
\begin{align}
  \label{eq:langevin}
  \lim_{N_\ell \rightarrow \infty} \frac{d \R_\ell}{dt} &= 
  \E\sqb{\left. \frac{d \R_\ell}{dt} \right| \R_\ell} = 
  \tfrac{1}{2 N_\ell} dt \b{\b{\dd[\mathcal{L}]{\V_\ell}} \V_\ell^T + \V_\ell \b{\dd[\mathcal{L}]{\V_\ell}}^T} + dt \I.
\end{align}
%And as we take the limit as $N_\ell \rightarrow \infty$, these updates become deterministic,
%\begin{align}
%  \lim_{N_\ell \rightarrow \infty} \frac{d \R_\ell}{dt} &= 
%  \tfrac{1}{2 N_\ell} \b{\b{\dd[\mathcal{L}]{\V_\ell}} \V_\ell^T + \V_\ell \b{\dd[\mathcal{L}]{\V_\ell}}^T} + dt \I.
%\end{align}
To check that these dynamics are sensible, we consider performing Langevin sampling using the above dynamics under the zero-mean, unit-variance prior on elements of $\V_\ell$,
%In particular, the prior on elements of $\V_\ell$ is an independent, zero-mean unit-variance Gaussian, so the objective is,
\begin{align}
  \label{eq:simple-L}
  \mathcal{L} &= -\tfrac{1}{2} \tr\b{\V_\ell \V_\ell^T},
  \intertext{so the gradient is,}
  \dd[\mathcal{L}]{\V_\ell} &= \dd{\V_\ell}\sqb{-\tfrac{1}{2} \tr\b{\V_\ell \V_\ell^T}} = -\V_\ell.
  \intertext{Thus,}
  \E\sqb{\left. \frac{d \R_\ell}{dt} \right| \R_\ell} &= 
  \tfrac{1}{N_\ell} \V_\ell \V_\ell^T + \I = -\R_\ell + \I.
\end{align}
Now, we set the expected change in $\R_\ell$ equal to zero,
\begin{align}
  \0 &= \E\sqb{\frac{d \R_\ell}{dt}} = -\E\sqb{\R_\ell} + \I.
\end{align}
and solving for the expected value of $\R_\ell$,
\begin{align}
  \E\sqb{\R_\ell} = \E\sqb{\tfrac{1}{N_\ell} \V_\ell \V^T_\ell} &= \I,
\end{align}
which is equal to the expected value of $\tfrac{1}{N_\ell} \V_\ell \V_\ell^T$ under the prior, as is necessary given that these dynamics perform exact Langevin sampling in the limit.

\subsection{Langevin dynamics as the modes of an objective}

We can write the expected dynamics of $\R_\ell$ under Langevin sampling as the gradient of a surrogate objective, $\mathcal{L}'$,
\begin{align}
  \label{eq:langevin:modified}
  \mathcal{L}' &= \mathcal{L} + \tfrac{N_\ell}{2} \log \abs{\R} = \mathcal{L} + \tfrac{N_\ell}{2} \log \abs{\V_\ell \V_\ell^T}.
\end{align}
The gradient of the determinant is given by the pseudo-inverse,
\begin{align}
  %\dd{\V_\ell^T} \log \abs{\tfrac{1}{N} \V_\ell \V_\ell^T} &= 
  %2 \b{\b{\V_\ell \V_\ell^T}^{-1} \V_\ell}^T
  \dd{\V_\ell} \log \abs{\tfrac{1}{N_\ell} \V_\ell \V_\ell^T} &= 
  2 \b{\V_\ell \V_\ell^T}^{-1} \V_\ell.
\end{align}
Thus, continuous gradient descent on the full objective, with a learning rate of $\tfrac{1}{2}$, gives,
\begin{align}
  d\V_\ell &= \tfrac{1}{2} dt \sqb{\dd[\mathcal{L}']{\V_\ell}} = \tfrac{1}{2} dt \sqb{\dd[\mathcal{L}]{\V_\ell} + N_\ell \b{\V_\ell \V_\ell^T}^{-1} \V_\ell} 
  %d \V_\ell^T &= \tfrac{1}{2} dt \b{\dd[\mathcal{L}]{\W} + N \W \b{\W^T \W}^{-1}}.
\end{align}
The implied change in $\R$ is,
\begin{align}
  d\R_\ell &= \tfrac{1}{N_\ell} \b{d\V_\ell \V_\ell^T + \V_\ell d\V_\ell^T} = \tfrac{dt}{2 N_\ell} \b{\b{\dd[\mathcal{L}]{\V_\ell}} \V_\ell^T + \V_\ell \b{\dd[\mathcal{L}]{\V_\ell}}^T} + dt \I
\end{align}
And this is exactly equal to the change in $\R$ induced by Langevin sampling Eq.~\eqref{eq:langevin}.

\subsection{The sampling objective as modified maximum-likelihood under a Wishart prior}
To further check that the Langevin sampling result is sensible, we note that it is very similar to doing MAP inference under a Wishart prior, but that sampling fixes pathologies in this proceedure due to the skew inherent in the Wishart distribution.

In particular, the Wishart probability density is given by,
\begin{align}
  \log \P{\K_\ell| \J_{\ell}} &= \log \operatorname{Wishart}\b{\K_\ell; \tfrac{1}{N_\ell} \J_{\ell}, N_\ell}\\
  &= \tfrac{N_\ell-P-1}{2} \log \abs{\K_\ell} - \tfrac{N_\ell}{2} \log \abs{\J_{\ell}}   - \tfrac{N_\ell}{2} \tr\b{{\J^{-1}_{\ell-1} \K_\ell}}.
\end{align}
The pathologies arise if we compare the expectation and the mode of this distribution,
\begin{subequations}
\begin{align}
  \E\sqb{\K_\ell| \J_\ell} &= \J_\ell\\
  \argmax_{\K_\ell} \sqb{\log \P{\K_\ell| \J_{\ell}}} &= \b{N_\ell - P - 1} \J_\ell,
\end{align}
\end{subequations}
where the matricies $\K_\ell$ and $\J_\ell$ are $P \times P$, and $\K_\ell$ is the inner product of $N_\ell$ vectors with covariance $\tfrac{1}{N_\ell} \J_\ell$.
Thus, the mode gives a very poor characterisation of the expectation of the distribution, to the extent if $N_\ell = P+1$, the mode is zero while the expectation can take on any value.
Thankfully, it is possible to find a closely related optimization problem that gives a good characterisation of the mean.
In particular, we need to incorporate a new term in the objective that counteracts the ``shrinkage'' induced by the skew in the Wishart, such that the mode of the new objective equals the expectation,
\begin{align}
  \argmax_{\K_\ell} \sqb{\log \P{\K_\ell| \J_{\ell}} + \tfrac{P + 1}{2} \log \abs{\K_\ell}} &= \J_\ell.
\end{align}
Critically, this term, $\tfrac{P+1}{2} \log \abs{\K_\ell}$ is almost entirely independent of the parameters (it depends only on the size, $P$), and the combined objective is equivalent to the objective for Langevin sampling, $\mathcal{L}'$,
\begin{align}
  \label{eq:wishart:modified}
  \log \P{\K_\ell| \J_{\ell}} + \tfrac{P + 1}{2} \log \abs{\K_\ell} 
  &= \tfrac{N_\ell}{2} \log \abs{\J^{-1}_{\ell} \K_\ell} - \tfrac{N_\ell}{2} \tr\b{\J^{-1}_{\ell} \K_\ell}\\
  &= \tfrac{N_\ell}{2} \log \abs{\U^{-T}_{\ell} \K_\ell \U^{-1}_{\ell}} - \tfrac{N_\ell}{2} \tr\b{\U^{-T}_{\ell} \K_\ell \U^{-1}_{\ell}}\\
  &= \tfrac{N_\ell}{2} \log \abs{\R_{\ell}} - \tfrac{N_\ell}{2} \tr\b{\R_\ell}.
\end{align}
if we consider a simple one-layer setup, with $\mathcal{L}$ given by Eq.~\eqref{eq:simple-L}.
%Remarkably, this modified objective is equal to the objective corresponding to Langevin sampling derived above, if we remember that 
%The mode of this modified objective is indeed equal to the expectation,

\subsection{Representation learning in deep networks}

The log-probability of the data at the final layer can be written in the same form as the objective for Langevin sampling (Eq.~\ref{eq:langevin:modified}), and the modified objective for Wishart inference (Eq.~\ref{eq:wishart:modified}).
In particular,
\begin{align}
  \log \P{\y_\mu| \K_L} &= -\tfrac{1}{2} \y_\mu^T \L_L^{-1} \y_\mu - \tfrac{1}{2} \abs{\K_L} + \const.
  \intertext{defining the constant kernel, $\K_{L+1} = \tfrac{1}{Y} \Y \Y^T$, we can write the log-probability of $\Y$ in a manner that is consistent with the previous kernels,}
  \log \P{\Y| \K_L} &= \tfrac{Y}{2}\b{\log \abs{\L^{-1}_L \K_{L+1}} -\tr\b{\L_L^{-1} \K_{L+1}}} + \const
\end{align}
where the log determinant of $\K_{L+1}$ is constant, so can be included without changing the objective.

%For a deep network, taking $\K_0 =\tfrac{1}{X} \X^T \X$, the objective for all but the last layer is given by,
%\begin{align}
%  \nonumber
%  \mathcal{L}\b{\K_1,\dotsc,\K_L| \K_0} &= \sum_{\ell=1}^{L} \mathcal{L}\b{\K_{\ell}| \K_{\ell-1}} = \tfrac{N}{2} \sum_{\ell=1}^{L} \b{\log \abs{\L_{\ell-1}^{-1} \K_\ell} - \tr\b{\L_{\ell-1}^{-1} \K_{\ell}}}.
%\end{align}
%The log-probability of the data given the kernel at the final layer, again ignoring additive constants, is
%\begin{align}
%  \log \P{\y_\mu| \K_L} &= -\tfrac{1}{2} \y_\mu^T \L_L^{-1} \y_\mu - \tfrac{1}{2} \abs{\K_L} + \const.
%  \intertext{defining the constant kernel, $\K_{L+1} = \tfrac{1}{Y} \Y \Y^T$, we can write the log-probability of $\Y$ in a manner that is consistent with the previous kernels,}
%  \mathcal{L}\b{\Y| \K_L} &= \tfrac{Y}{2}\b{\log \abs{\L^{-1}_L \K_{L+1}} -\tr\b{\L_L^{-1} \K_{L+1}}} + \const
%\end{align}
As such, the full objective can be written as,
\begin{align}
  \mathcal{L} &= \sum_{\ell=1}^{L+1} \tfrac{N_\ell}{2} \b{\log \abs{\L_{\ell-1}^{-1} \K_\ell} - \tr\b{\L_{\ell-1}^{-1} \K_{\ell}}}.
\end{align}
When we differentiate, only the terms that vary with $\K_\ell$ are relevant,
\begin{align}
  \mathcal{L} =& \tfrac{N_\ell}{2} \b{\log \abs{\L_{\ell-1}^{-1} \K_\ell} - \tr\b{\L_{\ell-1}^{-1} \K_{\ell}}} +\tfrac{N_{\ell+1}}{2} \b{\log \abs{\L_\ell^{-1} \K_{\ell+1}} - \tr\b{\L_\ell^{-1} \K_{\ell+1}}}.
\end{align}
While the derivations up to this point have been the same, the gradients are different for fully connected, locally connected, and convolutional networks diverge.

\subsection{Fully connected networks}
For fully connected networks,
\begin{align}
  \L_\ell &= \K_\ell.
\end{align}
so the terms in the objective that depend on $\K_\ell$ are,
\begin{align}
  \mathcal{L} =& \tfrac{N_\ell}{2} \b{\log \abs{\K_{\ell-1}^{-1} \K_\ell} - \tr\b{\K_{\ell-1}^{-1} \K_{\ell}}} +\tfrac{N_{\ell+1}}{2} \b{\log \abs{\K_\ell^{-1} \K_{\ell+1}} - \tr\b{\K_\ell^{-1} \K_{\ell+1}}}.
\end{align}
Differentiating the relevant terms,
\begin{subequations}
\begin{align}
  \dd[\tr{\K_\ell^{-1} \K_{\ell+1}}]{\K_\ell} &= -\K_\ell^{-1} \K_{\ell+1} \K_\ell^{-1}\\
  \dd[\tr{\K_{\ell-1}^{-1} \K_\ell}]{\K_\ell} &= \K_{\ell-1}^{-1}\\
  \dd[\log \abs{\K_{\ell-1}^{-1} \K_\ell}]{\K_\ell} &= \dd[\log \abs{\K_\ell}]{\K_\ell} = \K_\ell^{-1}\\
  \dd[\log \abs{\K^{-1}_\ell \K_{\ell+1}}]{\K_\ell} &= -\dd[\log \abs{\K_\ell}]{\K_\ell} = -\K_\ell^{-1}
\end{align}
\end{subequations}
We then set the gradients to zero,
\begin{align}
  %\label{eq:MAP:to-solve}
  \0 &= \dd[\mathcal{L}]{\K_\ell} = -\b{N_{\ell+1}-N_\ell} \K_\ell^{-1} + N_{\ell+1} \; \K_\ell^{-1} \K_{\ell+1} \K_\ell^{-1} - N_\ell \; \K_{\ell-1}^{-1}
\end{align}
We pre multiply by $\K_\ell$,
\begin{align}
  \0 &= -\b{N_{\ell+1}-N_\ell} \I + N_{\ell+1} \; \K_{\ell+1} \K_\ell^{-1} - N_\ell \; \K_\ell \K_{\ell-1}^{-1},
\end{align}
And note that the resulting expression can be written in terms of a ratio, $\T_{\ell+1} = \K_{\ell+1} \K_\ell^{-1}$
\begin{align}
  \0 &= -\b{N_{\ell+1}-N_\ell} \I + N_{\ell+1} \; \T_{\ell+1} - N_\ell \; \T_\ell.
\end{align}
Solving for $\T_{\ell+1}$,
\begin{align}
  \T_{\ell+1} &= \I + \tfrac{N_\ell}{N_{\ell+1}} \b{\T_\ell - \I}
  %\b{1-\tfrac{N_\ell}{N_{\ell+1}}} \I + \tfrac{N_\ell}{N_{\ell+1}} \R_\ell
\end{align}
We use $N_\ell = N$, for $\ell \in \{1,\dotsc,N\}$,and $N_{L+1} = Y$,
\begin{align}
  \T_{\ell} &= \begin{cases}
    \T                            & \text{for } \ell \in \{1,\dotsc,L\}\\
    \I + \tfrac{N}{Y} \b{\T - \I} & \text{for } \ell = L+1
  \end{cases}
\end{align}
to compute $\T$, we use,
\begin{align}
  \K_{L+1} \K_0^{-1} &= \T_{L+1} \T^L,
  \intertext{substituting for $\T_{L+1}$,}
  \K_{L+1} \K_0^{-1} &= \b{\I + \tfrac{N}{Y}\b{\T - \I}} \T^L
\end{align}
As this cannot be solved analytically for $\T$, we consider three special cases.
First, if there are many outputs in comparison to the number of hidden units (i.e.\ $N/Y \rightarrow 0$), 
\begin{align}
  \lim_{N/Y \rightarrow 0} \T &= \b{\K_{L+1} \K_0^{-1}}^{1/L}
\end{align}
and thus, the top-level kernel is equal to the output kernel, i.e.\ $\K_L = \K_{L+1}$.
Second, we consider the other extreme where there are many more hidden units than output channels (i.e.\ $N/Y \rightarrow \infty$).
In this limit, we must have $\0 = \T - \I$ because otherwise the $\tfrac{N}{Y} \b{\T - \I}$ term will explode,
\begin{align}
  \lim_{N/Y \rightarrow \infty} \T &= \I,
\end{align}
thus, the representation does not change as it flows throught the network.
Finally, we consider a more reasonable case where the number of hidden units is of the order of the number of output channels --- in particular, we consider $Y=N$,
\begin{align}
  \T &= \b{\K_{L+1} \K_0^{-1}}^{1/(L+1)}
  \intertext{as such, the top-layer kernel is almost --- but not quite --- equal to the output kernel, but it does get closer as the network gets deeper,}
  \K_L &= \T^L \K_0 = \b{\K_{L+1} \K_0^{-1}}^{L/(L+1)} \K_0
\end{align}

\section{Natural gradients for a Gaussian-process sum kernel}
\label{sec:app:natural-gradients}
We begin by defining the covariance (kernel) as the sum over a set of kernels, $\K_i$, weighted by $\lambda_i$,
\begin{align}
  \K &= \sum_i \lambda_i \K_i.
\end{align}
Our goal is to find the maximum-likelihood $\lambda_i$ parameters using a natural-gradient method.
The likelihood is,
\begin{align}
  \log \P{\Y} &= -\tfrac{1}{2} \tr\b{\K^{-1} \Y \Y^T} - \tfrac{N}{2} \log \abs{\K} + \const.
\end{align}
And the gradient is,
\newcommand{\La}{\L_\alpha}
\newcommand{\Lb}{\L_\beta}
\newcommand{\Ly}{\L_\text{y}}
\begin{align}
  \dd[\log \P{\Y}]{\lambda_\alpha} &= \tfrac{1}{2} \tr{\La \Ly} - \tfrac{N}{2} \tr{\La}.
  \intertext{where,}
  \La &= \K^{-1} \K_\alpha\\
  \Ly &= \K^{-1} \Y \Y^T
\end{align}

For a natural-gradient method, we need the expected-second-derivatives.
For the first term, these are,
\begin{align}
  \E\sqb{\frac{\partial}{\partial \lambda_\beta}\sqb{\tfrac{1}{2} \tr{\La \Ly}}}
  &= \E\sqb{- \tfrac{1}{2} \b{\tr{\Lb \La \Ly} + \tr{\La \Lb \Ly}}}\\
  &= - \tfrac{1}{2} \b{\tr{\Lb \La \E{\sqb{\Ly}}} + \tr{\La \Lb \E{\sqb{\Ly}}}}\\
  &= - \tfrac{N}{2} \b{\tr{\Lb \La} + \tr{\La \Lb}}\\
  &= - N \tr{\La \Lb}
\end{align}
using basic matrix identities, and the fact that, under the model, $\E{\sqb{\Ly}} = N \I$.
The second term is independent of $\Y$, so we can just compute the second derivative,
\begin{align}
  \nonumber
  \frac{\partial}{\partial \lambda_\beta} \sqb{-\tfrac{N}{2} \tr{\La}}
  &= \tfrac{N}{2} \tr{\Lb \La}.
\end{align}
Thus,
\begin{align}
  \E\sqb{\frac{\partial^2}{\partial \lambda_\alpha \lambda_\beta} \log \P{\Y}} &= -\tfrac{N}{2} \tr{\La \Lb}
\end{align}

\end{document}